\PassOptionsToPackage{table}{xcolor}

\pdfoutput=1

\documentclass[11pt]{article}

\usepackage[final]{acl}

\usepackage{times}
\usepackage{latexsym}

\usepackage[T1]{fontenc}

\usepackage[utf8]{inputenc}

\usepackage{microtype}

\usepackage{inconsolata}

\usepackage{graphicx}

\usepackage{blindtext}
\usepackage{amsmath}
\usepackage{listings}
\usepackage{multirow}
\usepackage{arydshln}
\usepackage{booktabs}
\usepackage{array}
\usepackage{color}
\usepackage{colortbl}
\usepackage{xcolor}
\usepackage{balance}

\usepackage{tabularx}
\usepackage{adjustbox}
\usepackage{makecell}
\usepackage{hyperref}
\usepackage{tcolorbox}
\tcbuselibrary{breakable}
\let\svthefootnote\thefootnote
\usepackage{wrapfig}     
\usepackage{pifont}

\newcommand\freefootnote[2]{%
  \let\thefootnote\relax%
  \footnotetext{\begin{minipage}[t]{\textwidth}%
    \raggedright
    #1 \\ #2
    \end{minipage}}%
  \let\thefootnote\svthefootnote%
}

\usepackage[bottom]{footmisc}

\usepackage{cuted}

\renewcommand{\footnotesize}{\small}

\usepackage{pgf}
\makeatletter
\newcommand{\groupedRowColors}[5][0]{
    \global\rownum=\z@
    \global\@rowcolorstrue
    \@ifxempty{#4}%
        {\def\@oddrowcolor{\@norowcolor}}%
        {\def\@oddrowcolor{\gdef\CT@row@color{\CT@color{#4}}}}%
    \@ifxempty{#5}%
        {\def\@evenrowcolor{\@norowcolor}}%
        {\def\@evenrowcolor{\gdef\CT@row@color{\CT@color{#5}}}}%
    \def\@rowcolors{%
        \if@rowcolors
            \noalign{%
                \relax
                \ifnum\rownum<#3
                    \@norowcolor
                \else \ifodd \numexpr (\rownum - (\rownum/#2)*(#2))\relax
                    \@oddrowcolor
                \else
                    \@evenrowcolor
                \fi \fi
            }%
        \fi
    }%
    \CT@everycr{\@rowc@lors\the\everycr}%
    \ignorespaces
}
\makeatother

\title{RAD-Bench: Evaluating Large Language Models' Capabilities in Retrieval Augmented Dialogues}

\author{
 $^{1,2}$ \textbf{Tzu-Lin Kuo\textsuperscript{\dag*}},
 $^{2}$ \textbf{Feng-Ting Liao} \textsuperscript{*},
 $^{2}$ \textbf{Mu-Wei Hsieh}\textsuperscript{\dag},\\
 $^{2}$ \textbf{Fu-Chieh Chang},
 $^{2}$ \textbf{Po-Chun Hsu},
 $^{2}$ \textbf{Da-Shan Shiu}
 \\
 \\
 $^{1}$ National Taiwan University, $^{2}$ MediaTek Research
\\
\texttt{r12922050@ntu.edu.tw, \{ft.liao, morris-mw.hsieh,}\\
\texttt{mark-fc.chang, pochun.hsu, ds.shiu\}@mtkresearch.com}
}

\begin{document}
\maketitle
\begin{abstract}
In real-world applications with Large Language Models (LLMs), external retrieval mechanisms—such as Search-Augmented Generation (SAG), tool utilization, and Retrieval-Augmented Generation (RAG)—are often employed to enhance the quality of augmented generations in dialogues. These approaches often come with multi-turn dialogue, where each interaction is enriched by relevant information retrieved from external sources. Existing benchmarks either assess LLMs' chat abilities in multi-turn dialogues or their use of retrieval for augmented responses in limited tasks such as knowledge QA or numeric reasoning. To address this gap, we introduce \textbf{RAD}-Bench (\textbf{R}etrieval \textbf{A}ugmented \textbf{D}ialogue), a  comprehensive benchmark designed to evaluate LLMs' capabilities in multi-turn dialogues following retrievals. RAD-Bench evaluates two key abilities of LLMs: \emph{Retrieval Synthesis} and \emph{Retrieval Reasoning} over 6 representative scenarios, concluded from analysis of real-world tasks. By employing discriminative questions, retrieved contexts, and reference answers, our evaluation of prevalent LLMs reveals performance degradation as additional layers of conditions or constraints are applied across conversation turns, even when accurate retrieved contexts are provided. The data and code are available at \url{https://github.com/mtkresearch/RAD-Bench}
\end{abstract}

\freefootnote{\textsuperscript{\dag}Work done during internship at MediaTek Research.}{\textsuperscript{*} Equal contribution.}

\section{Introduction}
\begin{figure}[t]
    \centering
    \includegraphics[width=\columnwidth]{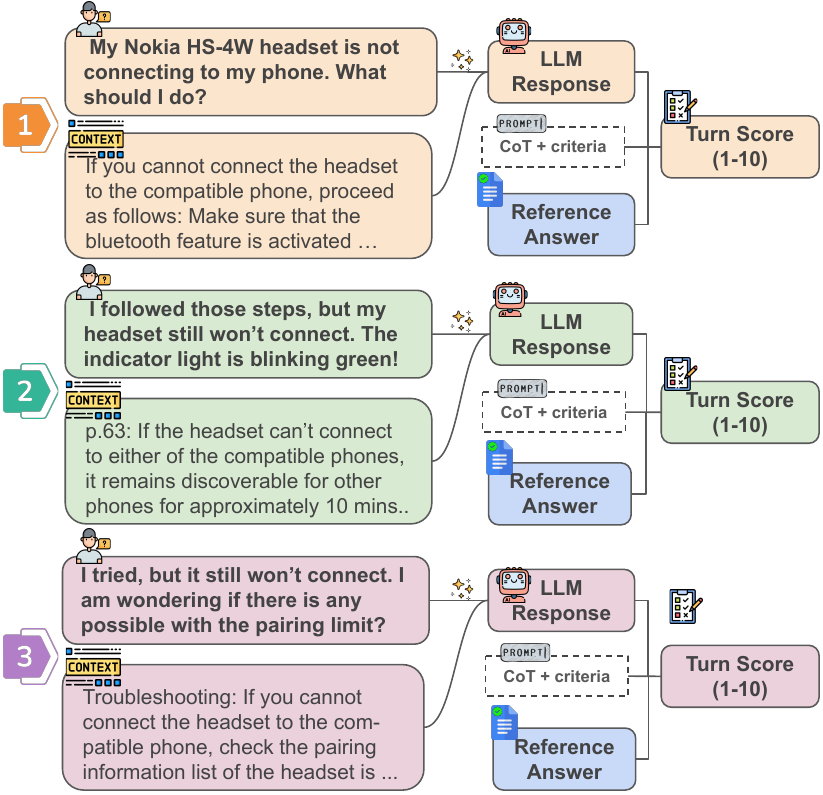}
    \caption{\textbf{Evaluation Process in Retrieval Augmented Dialogue Benchmark}: At each turn, a user question paired with a retrieved context is presented to the LLM for augmented generation. The LLM's response is scored on a scale of 1 to 10 using an LLM-as-a-Judge framework. This framework prompts the judge to assess how well the model utilized the given context to answer progressively changing questions, based on specific criteria, and compare it against a reference answer, ensuring accurate and consistent evaluations across different scenarios.}
    \label{fig:rad_teaser}
\vspace{-5pt}
\end{figure}


\begin{table*}[t]
    
\resizebox{\textwidth}{!}{
\begin{tabular}{lc|ccc:cc:c}
 & & \textbf{MultiDoc2Dial} & \textbf{ORConvQA} & \textbf{ConvFinQA} & \textbf{MT-Bench} & \textbf{Wild-Bench} & \textbf{RAD-Bench} \\
 & & \citet{feng-etal-2021-multidoc2dial} & \citet{chen-ORConvQA-2020} & \citet{chen-etal-2022-convfinqa} & \citet{zheng_judging_2023} & \citet{lin_wildbench_2024} & \textbf{(ours)}\\
\cmidrule{2-8}

\multirow{2}{*}{\rotatebox{90}{\textbf{Mode}}}&\textbf{Context Conditioning} & \ding{51} & \ding{51} & \ding{51} & \ding{55} & \ding{55} & \ding{51} \\
& \textbf{Multi-turn Questions} & \ding{51} & \ding{51} & \ding{51} & \ding{51} & \ding{55} & \ding{51}  \\

\cmidrule{2-8}

\multirow{3}{*}{\rotatebox{90}{\textbf{Stats.}}}& \textbf{Number of Tasks} &  1 & 1 & 2 & 8 & 11 & 6 \\
& \textbf{Question Turns} & >2 & >2 & >2 & 2 & 1 & 3 \\
& \textbf{Evaluated Samples} & 4796 & 5571 & 14115 & 160 & 1024 & 267 \\

\cmidrule{2-8}

\multirow{4}{*}{\rotatebox{90}{\textbf{Tasks}}}& \textbf{Knowledge QA} & \ding{51} & \ding{51} & \ding{51} & \ding{51} & \ding{51} & \ding{51} \\
& \textbf{Knowledge Summarization} & \ding{55} & \ding{55} & \ding{55} & \ding{51} & \ding{51} & \ding{51} \\
& \textbf{Chain of Reasoning} & \ding{55} & \ding{55} & \ding{51} & \ding{51}  & \ding{51} & \ding{51} \\
& \textbf{Planning} & \ding{55} & \ding{55} & \ding{55} & \ding{55} & \ding{51} & \ding{51} \\
\cmidrule{2-8}
\end{tabular}
}

    \caption{\textbf{A comparison of selected question answering datasets}. Dialogue and chat benchmarks typically cover the following key tasks: \textit{Knowledge QA}, involving factual question answering with factoids embedded in provided context; \textit{Knowledge Summarization}, requiring summarizing a context according to instructions; \textit{Chain of Reasoning}, centering on arithmetic reasoning with factoids resting within a context; and \textit{Planning}, involving following instructions to make plans using context with graph data structure. In RAD-Bench, scenarios in Retrieval Synthesis covers Knowledge QA and Knowledge summarization, while that in Retrieval Reasoning includes Knowledge QA, Chain of Reasoning, and Planning.
    }
    \label{tab:benchmark_comparison}
\end{table*}


In recent years, Large Language Models (LLMs) have demonstrated exceptional language understanding ability and have been applied across various industries, serving as assistants in fields such as academia, customer support, and research. \citep{kalla2023study}. Despite recent advances, LLMs still face challenges such as hallucination and inherent biases \citep{xu_hallucination_2024}. To address these issues without the high costs of retraining, many real-world applications \citep{openai_chatgpt_2023,mediatek_mediatek_2024,perplexity_ai_perplexity_2024} now utilize RAG \citep{lewis_retrieval-augmented_2020} to augment LLM outputs with retrieved context. This approach, which includes incorporating retrieved documents, web search results \citep{luo_search_2023}, and knowledge graphs \citep{xie_travelplanner_2024}, has become a common practice to enhance accuracy and reduce hallucination in LLM-generated content. With the growing reliance on retrieval-augmented LLMs in practical applications, there is an urgent need for a comprehensive benchmark that evaluates their ability to effectively utilize provided context.

Existing benchmarks for evaluating LLMs' augmented generation following retrieved context, such as \citet{lyu_crud-rag_2024}, \citet{chen_benchmarking_2024}, \citet{yang_crag_2024}, \citet{xie_travelplanner_2024}, and \citet{zheng_natural_2024}, focus on single-turn instructions, whereas real-world interactions involve multi-turn dialogues. Meanwhile, benchmarks in evaluating LLMs' chat capabilities in multi-turn dialogues, such as \citet{finch_dont_2022}, \citet{zheng_judging_2023}, and \citet{bai_mt-bench-101_2024}, neglect instruction-following with retrieved context. While goal-oriented dialogue research \citep{dinan2018wizard,feng-etal-2021-multidoc2dial} addresses multi-turn interactions with retrieved context, it often emphasizes factual grounding over comprehensive context generation quality for evolving queries in typical real-world scenarios such as writing, summarizing, and planning.

To address the aforementioned gap, we propose Retrieval Augmented Dialogue Benchmark (RAD-Bench), a benchmark designed to measure LLMs' ability to follow user instructions in multi-turn dialogue scenarios and effectively recall and utilize retrieved context to enhance their responses. Specifically, as shown in Figure \ref{fig:rad_teaser}, each benchmark sample consists of three-turn questions with accompanied retrieved context at each turn. RAD-Bench evaluates two key abilities of LLMs in multi-turn dialogues: \emph{Retrieval Synthesis} and \emph{Retrieval Reasoning}. These abilities are assessed through scenarios curated from real-world dialogue data \citep{dom_eccleston_sharegpt_2024,mediatek_mediatek_2024}. \textbf{Retrieval Synthesis} measures an LLM's ability to progressively integrate retrieved context for tasks like summarization and article writing, enabling effective knowledge accumulation and synthesis. \textbf{Retrieval Reasoning} evaluates whether LLM can make reasonable inference when user intent changes or additional conditions are introduced across turns, utilizing context in each turn to refine and improve responses. For each ability, we select three representative scenarios that exemplify multi-turn dialogues following retrievals. To construct RAD-Bench, we developed a pipeline leveraging multiple LLMs to generate, select, and synthesize questions and retrieved contexts, ensuring diverse, relevant, and high-quality benchmark samples through automated scoring and manual inspection. In total, RAD-Bench comprises 89 multi-turn question samples, each consisting of 3 turns with accompanying retrieved context and reference answer. This results in a total of 267 turns for evaluation.

To evaluate RAD-Bench, we employ the LLM-as-a-Judge framework \citep{zheng_judging_2023}, using scenario-specific criteria inspired by \citet{fu_gptscore_2023} as scoring guidelines. Our analysis includes both 4 closed-source and 8 open-source LLMs commonly used in industry. Results indicate a decline in model performance when new intents or conditions are introduced into multi-turn instructions, even when relevant retrieved contexts are provided. Additionally, by comparing the evaluation scores with Elo ratings from Chatbot Arena (Hard Prompts) \citep{li_crowdsourced_2024,chiang_chatbot_2024,tianle_li_live_2024}, we demonstrate that RAD-Bench effectively differentiates LLMs in context-rich, augmented dialogue applications. This comparison reveals that models with similar performance in standard multi-turn conversations may not maintain that performance in retrieval-augmented dialogues.

\begin{figure}[t]
  \includegraphics[width=\columnwidth]{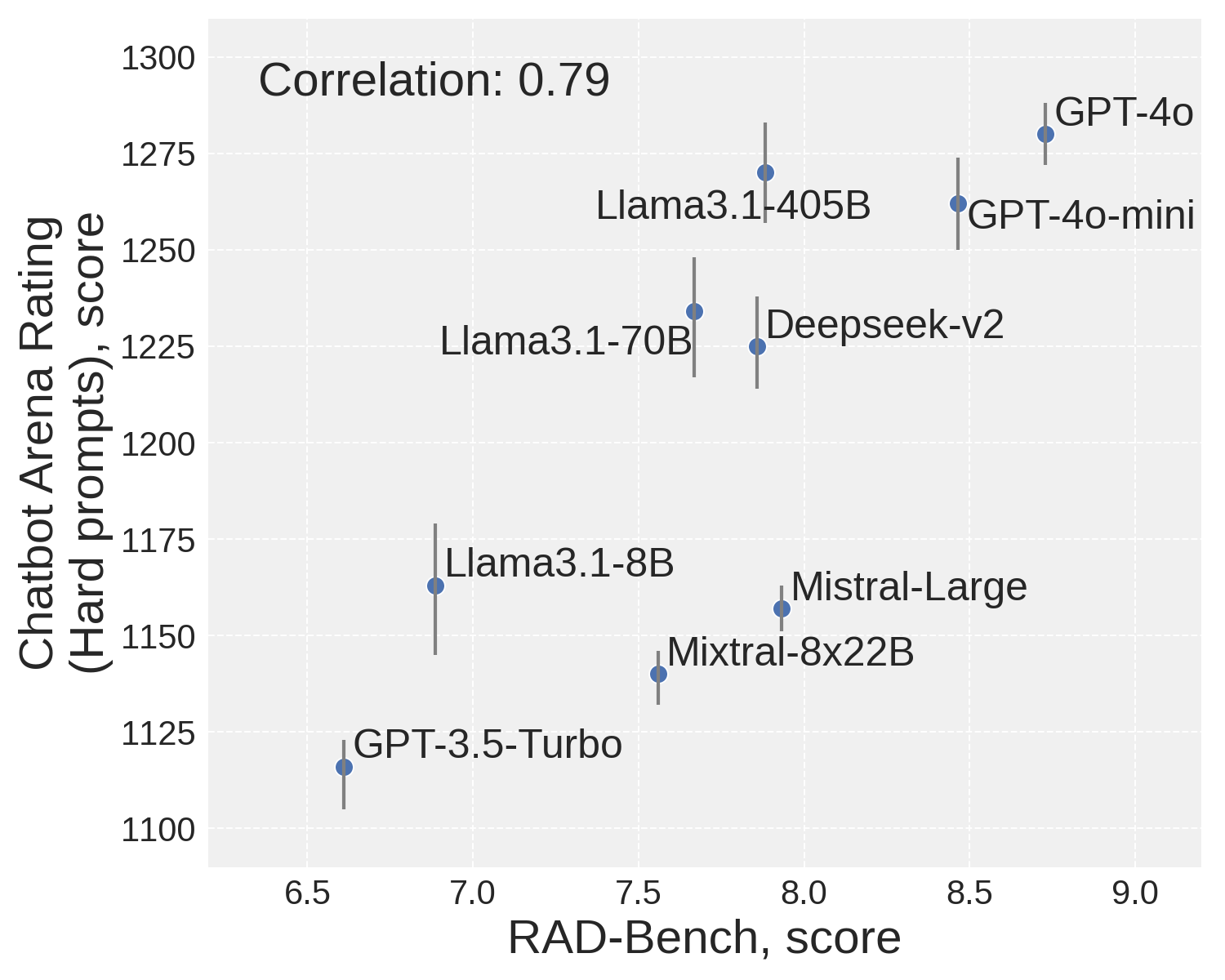}
  \caption{\textbf{Correlation between RAD-Bench and Chatbot Arena (Hard-En prompts)} \citep{chiang_chatbot_2024}. Models exhibiting similar level of multi-turn chat capability do not perform similarly when they are applied to dialogues from retrieval, as showcased by results from \texttt{Llama3.1-8B} vs \texttt{Mistral-Large}; from \texttt{Llama3.1-70B} vs \texttt{Deepseek-V2}; from \texttt{Llama3.1-405B} vs \texttt{GPT-4o}. We surmise that the discrepancy could be reduced through including RAFT \cite{zhang_raft_2024} in post-trainings, aligning model behaviors closer to the scenarios in retrieval augmented dialogue.}
  \label{fig:spearman_correlation}
\end{figure}
\vspace{-5pt}

\section{Related Work}

\subsection*{Retrieval Augmented Generation Benchmarks}

Several research efforts have evaluated LLMs' augmented generation ability with retrieved context. For instance, \citet{lyu_crud-rag_2024} evaluates RAG applications in \texttt{Create, Read, Update, and Delete} scenarios, while \citet{chen_benchmarking_2024} measures the fundamental abilities of LLMs required for RAG. Additionally, \citet{yang_crag_2024} comprehensively evaluate factual questions with context from documents, web searches, APIs, and knowledge graphs. Contexts from tools such as \texttt{Google Calendar} and \texttt{FlightSearch} are provided by \citet{xie_travelplanner_2024} and \citet{zheng_natural_2024} to LLMs for evaluating planning abilities. These benchmarks, though, evaluate LLMs in single-turn instructions, whereas real-world applications often involve multi-turn dialogues to address accumulation of hypotheses, constraints, and evolving user intents, which are not captured in typical single-turn evaluations. 

\subsection*{Context Grounded Dialogue Benchmarks}

To evaluate LLMs' ability to accurately adhere to instructions in multi-turn dialogues grounded on context in open-ended tasks, several benchmarks have been proposed. Early work in document-grounded dialogue by \citet{dinan2018wizard,feng-etal-2021-multidoc2dial} asses conversation agents' capability to utilize context from documents for answering factual questions. Work by \citet{chen-etal-2022-convfinqa} explores the chain of numerical reasoning of LLMs in conversational question answering on financial reports. Notably, \citet{chen-ORConvQA-2020} benchmarks the retrieved passages for multi-turn questions but miss the nuance in benchmarking engagement or understandability \citep{fu_gptscore_2023} of the generated text. These existing work are primarily focused on multi-turn factual inquiries or numerical arithmetic tasks for evaluating conversational LLMs.

Furthermore, recent work by \citet{zheng_judging_2023} evaluates models across core abilities such as writing, extraction, and reasoning with LLM-as-a-Judge, while \citet{bai_mt-bench-101_2024} proposes fine-grained assessments of real-life dialogues. \citet{dubois_length-controlled_2024} and \citet{lin_wildbench_2024} comprehensively evaluate models with human-chatbot conversation logs, though these are limited to single-turn instructions. While these studies address the effectiveness of LLMs in complex tasks like knowledge synthesis, summarization, planning, and reasoning, they often overlook the aspect of context retrieval, which is crucial for applications rich in contextual information.

To bridge this gap, we propose RAD-Bench for a comprehensive evaluation of common knowledge synthesis and reasoning tasks under retrieval augmented dialogues. \autoref{tab:benchmark_comparison} presents the comparison of our benchmark with existing ones.

\section{Retrieval Augmented Dialogue Benchmark}

As illustrated in \autoref{fig:rad_teaser}, each benchmark sample in RAD-Bench consists of three-turn questions with accompanied retrieved context to simulate the retrieval augmented dialogues. Responses to the turn questions by an LLM are evaluated by a reference-guided-judge, and a point-wise evaluation score for the LLM is reported. In the following section, we first introduce the two evaluated abilities in the benchmark: \emph{Retrieval Synthesis} and \emph{Retrieval Reasoning}, where each ability comes with three representative tasks, concluded through analysis of chat dialogues from ShareGPT \citep{dom_eccleston_sharegpt_2024}, and MediaTek DaVinci \citep{mediatek_mediatek_2024}. We then explain the reference-guided-judge for evaluating LLM in generating response for retrieval augmented dialogues and the construction pipeline of the benchmark.

    

\subsection{Evaluated Abilities}


\subsubsection*{Retrieval Synthesis}
We define \emph{Retrieval Synthesis} (RS) as the ability of LLM in following user instructions across turns while extracting useful information from retrieved information and integrating the information progressively. In the applications of RAG and SAG in chatbots \citep{perplexity_ai_perplexity_2024,mediatek_mediatek_2024}, users can require LLMs to utilize retrieved context for answering queries related to completing tasks such as summarization, paragraph writing, and knowledge synthesis in multi-turn dialogues. To measure the capability of LLMs in completing such tasks, we selected the following scenarios:

\begin{itemize}
\item \textbf{News TLDR (Too Long; Didn't Read)} embodies the scenario of journalist writing articles. It consists of instructions requiring LLMs to write comprehensive news articles by integrating retrievals of related past events, statistics, expert opinions, and recent developments. 

\item \textbf{Education} represents the case where educators compose educational articles. It comprises queries instructing LLMs to create engaging materials with progressive depths and breadths from retrievals of diverse educational resources.

\item \textbf{Academic Writing} exemplifies the scenario that researchers leveraging LLMs to draft and refine sections such as related work and literature reviews for academic papers. It includes multi-turn prompts that guide LLMs to integrate retrieved information from relevant studies, data, and citations, progressively building content depth.

\end{itemize}

\subsubsection*{Retrieval Reasoning}
We define \emph{Retrieval Reasoning} (RR), an ability of LLMs in adjusting responses using retrieved references to support logical reasoning and problem-solving across multiple dialogue turns with progressive change of conditions and constraints. Reasoning tasks such as data analysis \citep{mediatek_mediatek_2024}, constructing customer support chatbots \citep{pandya_automating_2023}, or planning \citep{xie_travelplanner_2024} through utilizing external databases and RAG are prevalent scenarios for LLM applications. In these scenarios, users interact with LLMs through queries that involve diverse hypotheses, new conditions, or changing intents based on retrieved information. We select scenarios where understanding context and evolving conditions is crucial for measuring the RR ability of LLMs. These are:

\begin{itemize}
\item \textbf{Customer Support} addresses the application of RAG techniques with LLMs to enhance the user experience of customer support chatbots. It consists of questions and retrieved contexts for evaluating LLMs in resolving customer inquiries and narrowing down solutions with the contexts as customers describe issues in more details progressively. 

\item \textbf{Finance} exemplifies the task of financial analyst utilizing LLMs with RAG to carry out data analysis. Queries in this scenario include tasks such as comparison of assets and computing finance metrics from retrieved financial statements for consolidating financial outlooks of companies at the end of multi-turn dialogues.

\item \textbf{Travel Planning} represents the case where LLMs act as travel planning assistants in suggesting travel itineraries based on external databases. Instructions in such scenario start from broad questions and move on to specific conditions, e.g., preferred destinations, budgets, accommodations, and activities, to test LLMs in reasoning through conditions with retreived contexts. Furthermore, conflicting and updates to conditions are presented in the multi-turn instructions to evaluate LLMs ability in correcting its advice. 

\end{itemize}

\begin{table*}[h]


\resizebox{\textwidth}{!}{
\begin{tabular}{c|c|cc|rrr|rrr|r}
    \toprule
        \multicolumn{4}{c|}{\textbf{Model}} & \multicolumn{7}{c}{\textbf{RAD-Bench}} \\
        \cmidrule{1-11}
        \textbf{Type} & \textbf{Name} & \makecell{\textbf{Activated} \\ \textbf{Params.}} & \makecell{\textbf{Context} \\ \textbf{Length}} & \textbf{Academic} & \textbf{News} & \textbf{Education} & \textbf{Finance} & \textbf{Customer} & \textbf{Travel} & \textbf{Average} \\


        
        \midrule
        \multirow{3}{*}{\rotatebox{90}{\textbf{Close}}} & GPT-4o & - & 128k & \underline{8.77} & \underline{8.68} & \underline{8.95} & \underline{9.00} & \underline{9.10} & 7.83 & \underline{8.72} \\
        & GPT-4o-mini & - & 128k & 8.27 & 8.53 & 8.80 & 8.87 & 8.53 & 7.80 & 8.47 \\
        & Mistral-Large & - & 32k & 8.17 & 7.77 & 8.33 & 8.58 & 7.83 & 6.76 & 7.91 \\
        & GPT-3.5-Turbo & - & 16k & 5.30 & 5.23 & 6.55 & 8.04 & 8.47 & 5.93 & 6.59 \\
        \midrule
        \multirow{8}{*}{\rotatebox{90}{\textbf{Open}}} 
        & Llama3.1-405B & 405B & 128k & 7.90 & 8.07 & 8.25 & 8.22 & 7.63 & 7.21 & \textbf{7.88} \\
        & Llama3.1-70B & 70B & 128k & 8.03 & 7.72 & 8.25 & 8.02 & 6.83 & 7.07 & 7.65 \\
        & Mixtral-8x22b & 39B & 64k & 7.70 & 7.47 & 7.97 & 8.22 & 8.10 & 5.79 & 7.54 \\
        & Deepseek-v2 & 21B & 128k & 7.57 & 6.67 & 8.00 & \textbf{8.71} & \textbf{8.27} & \underline{\textbf{7.95}} & 7.86 \\
        & BreeXe-8x7B & 13B & 8k & \textbf{8.47} & \textbf{8.14} & \textbf{8.58} & 7.56 & 7.63 & 5.74 & 7.69 \\
        & Mistral-Nemo-12B & 12B & 128k & 7.20 & 6.84 & 7.42 & 7.33 & 7.47 & 3.55 & 6.63 \\
        & Llama3.1-8B & 8B & 128k & 7.33 & 6.16 & 7.53 & 8.33 & 6.77 & 5.17 & 6.88 \\
        & Breeze-7B & 7B & 8k & 7.47 & 7.33 & 7.80 & 6.93 & 7.13 & 4.83 & 6.92 \\

    \bottomrule
\end{tabular}
}
    \caption{Evaluated models in RAD-Bench. For each scenario, \textbf{bold score} indicates the best open-weight model; \underline{underlined score} marks the best model overall. We report instruct versions of the open-weight models.}
    
    \label{tab:main_results}
\end{table*}

\subsection{Evaluator}
Trained with Reinforcement Learning from Human Feedback (RLHF), LLMs have demonstrated strong alignment with human preferences \citep{zheng_judging_2023}, achieving evaluation performance comparable to human experts \citep{bai_mt-bench-101_2024} while significantly reducing costs and improving scalability in model evaluation. Following \citet{zheng_judging_2023, fu_gptscore_2023, liu_g-eval_2023, bai_mt-bench-101_2024}, we utilize LLM-as-a-Judge and prompt the judge to evaluate chatbot responses to benchmark questions. The judge takes in chat history, retrieved context, and current turn question and response as inputs and provide a point-wise score to model response for each turn. Inspired by \citet{fu_gptscore_2023}, we devise evaluation criteria for judge prompts. Each criterion is accompanied by tailored instructions to guide the LLM's evaluation. For Retrieval Synthesis, we assess Consistency, Informativeness, and Coherence, while for Retrieval Reasoning, we evaluate Accuracy, Consistency, and Coherence. We implemented reference-guided judges \cite{zheng_judging_2023} with audited reference answers (Appendix \ref{subsec:reference_answer}) for each turn and adopt chain-of-thought to generate analysis based on the criteria and the reference answer before producing the final score. For further details of the judge prompts and definitions of above criteria, see Appendix \ref{appendix:promopts_for_eval}.

\subsection{Benchmark Construction}

\label{subsec:benchmark_construction}

To construct benchmark questions with auditable reference answers, we propose a data generation pipeline (Figure \ref{fig:rad_data_construction_pipeline}) that generates questions synthetically. This process involves deconstructing the knowledge points of an article into multiple-turn questions for Retrieval Synthesis and breaking down the joint conditions of solved tasks into multiple-turn questions for Retrieval Reasoning. We leverage LLMs both as question generators to create a pool of synthetic candidates and as question scorers to select the most suitable synthetic candidates for multi-turn dialogues from the retrievals. Detailed explanations of each phase are provided in Appendix \ref{sec:details_on_data_generation}.

\section{Evaluation Results}
\label{subsec:evaluation_results}

\subsection{Evaluation Setup}

We evaluated a series of models, including OpenAI GPT \citep{openai_chatgpt_2023}, Mistral \citep{jiang_mistral_2023}, Gemma \citep{gemma_team_gemma_2024}, Llama \citep{dubey_llama_2024}, DeepSeek \citep{deepseek-ai_deepseek-v2_2024}, and BreeXe \citep{hsu_breeze-7b_2024}, each available in multiple model sizes. All selected models have context windows more than 8k, suitable for RAD applications. Responses from closed-source models were collected in July 2024 and evaluated using \texttt{GPT-4o (2024-05-13)} with temperature set to 0.

\subsection{Main Results}
We show scores of evaluated models in Table \ref{tab:main_results} and in Figure \ref{fig:model_result_radar}. Overall speaking, the closed-source models, particularly \texttt{GPT-4o} with average of 8.72, consistently outperformed the open-source models across most scenarios. As for the open-source models, \texttt{Llama3.1-405B} and \texttt{Deepseek-v2} show strong performance with averages of 7.88 and 7.86, respectively. These two models stand out within the open-source category, though still trailing behind the top closed-source models.

\subsubsection*{Scenario-Specific Observations}
In Retrieval Synthesis scenarios, \texttt{BreeXe-8x-7B} achieved impressive performance, closely rivaling \texttt{GPT-4o-mini} and \texttt{GPT-4o}. This may due to \texttt{BreeXe-8x-7B}'s role as a question scorer, potentially biasing question selection towards its strengths. Additionally, Travel Planning scenario emerged as the most challenging, with \texttt{Deepseek-v2} outperforming all other models, including  \texttt{GPT-4o}. We attribute \texttt{Deepseek-v2}'s success to its two-stage reinforcement learning (RL) training strategy \citep{deepseek-ai_deepseek-v2_2024}, which enhances reasoning capabilities through initial optimization on code and math tasks, followed by safety alignment adjustments. The similarity between travel planning and coding/math tasks in hypothesis formation and constraint modification likely contributed to Deepseek-v2's superior performance in this scenario.

\subsubsection*{Effect of Model Size}

For open-source models such as Llama3.1, Mistral, and Breeze, it is evident that as the model size increases, there is a notable improvement in reasoning capabilities, with the most significant growth observed in the Travel Planning scenario. This observation aligns with findings of \citet{bai_mt-bench-101_2024} and \citet{mondorf2024comparing}, which emphasize that as model scale increases, the model's ability to reason, employ strategies, and interact becomes more pronounced. See Figure \ref{fig:combined_radar_charts} for further illustration of the performance distribution of various model series.

\vspace{-5pt} 
\begin{figure}[htbp]
  \includegraphics[width=0.88\columnwidth]{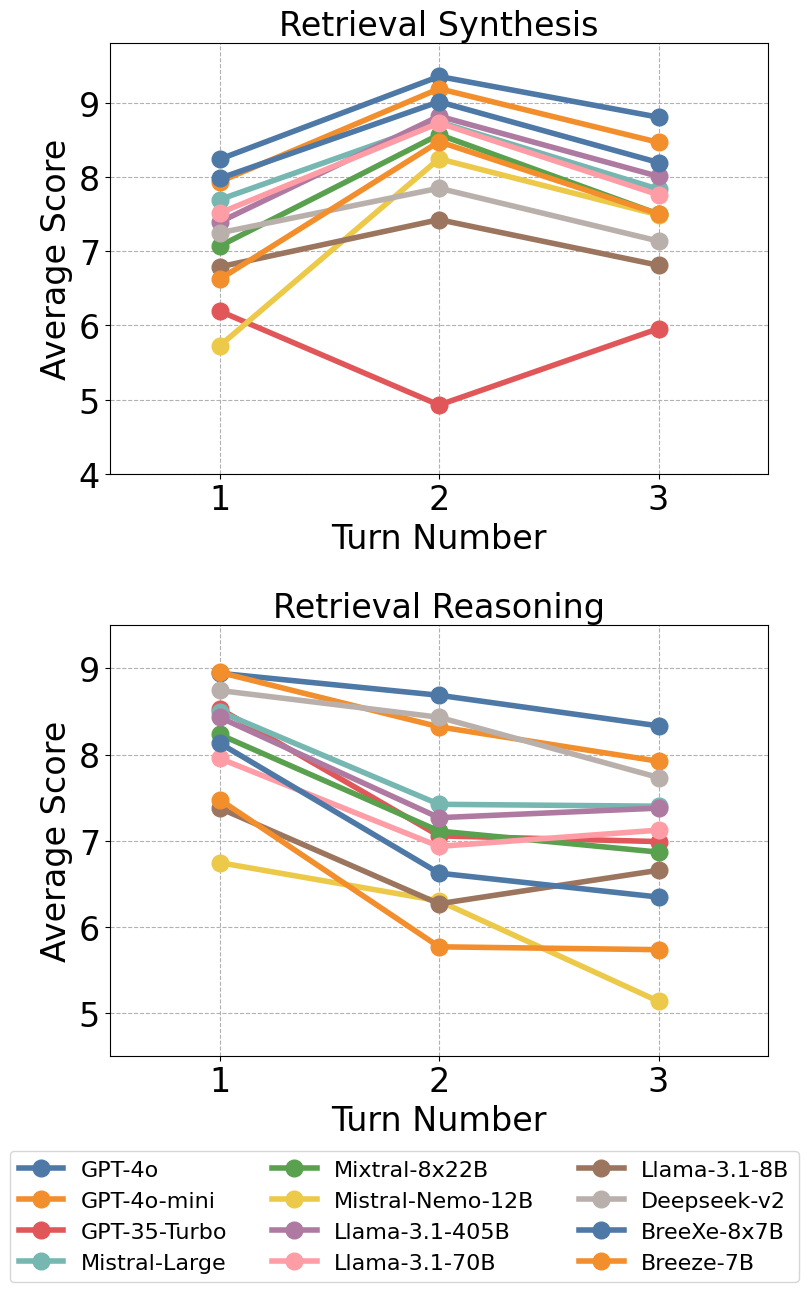}
  \caption{\textbf{Model performance across turns}. (Top): Retrieval Synthesis; (Bottom) Retrieval Reasoning.}
  \label{fig:scores_per_turn}
\end{figure}
\vspace{-10pt}

\subsection{Performance Across Dialogue Turns}

To investigate model performance across turns for different evaluated abilities, we calculate the average score for each dialogue turn, as shown in Figure \ref{fig:scores_per_turn}. In Retrieval Synthesis, model performance generally improves in the second turn but declines in the third. After carefully reviewing evaluator judgments, we attribute this to the nature of synthesis scenarios: second-turn questions typically extend the first turn's topic. Evaluators tend to give favorable scores as long as the response adheres to the general direction established in the first turn. As for the final turn, which requires summarizing diverse perspectives from previous rounds, presents a more complex task. For Retrieval Reasoning, scores decline with each turn. This is understandable, as new conditions or constraints in subsequent turns require more complex reasoning from the model, resulting in lower scores.

\subsection{Correlation with Chatbot Arena}
To study whether industry chat benchmark is sufficient to represent the performance of LLMs in applications requiring augmented generations, we compare the evaluation results of models in the benchmark to Elo scores of models from Chatbot Arena, an industry benchmark for assessing LLMs' chat capability \citep{chiang_chatbot_2024} through anonymous human evaluations. We include models appearing in the Chatbot Arena for comparison. Results in Figure \ref{fig:spearman_correlation} shows that RAD-Bench is discriminative. Models exhibiting similar level of chat capability, such as \texttt{GPT-4o} vs \texttt{Llama3.1-405B}; \texttt{Llama3.1-70B} vs \texttt{Deepseek-v2}; \texttt{Llama3.1-8B} vs \texttt{Mistral-Large}, do not perform equally well when the models are applied to scenarios with dialogues from retrieval. 

\section{Conclusions and Future Work}
\label{sec:discussion-future-work}
RAD-Bench provides significant value for industry applications by offering a comprehensive evaluation framework that assesses models' capabilities in augmented generation with retrieved context in multi-turn scenarios. By assessing both Retrieval Synthesis and Retrieval Reasoning across six practical scenarios inspired by human-LLM multi-turn dialogue interactions requiring retrieved context to complete tasks, RAD-Bench effectively differentiate model performance—even among LLMs with similar chat capabilities. This distinction is valuable for industries deploying retrieval-augmented LLM applications, as it demonstrates that traditional QA benchmarks and single-turn RAG benchmarks often fail to capture a model's effectiveness in these complex scenarios. By utilizing RAD-Bench, it helps companies optimize their model selection and deployment strategies, potentially saving significant resources while ensuring better performance in applications requiring multi-turn synthesis and reasoning with retrieved context.

In future work, expanding the diversity of questions and scenarios within RAD-Bench is crucial. While the current benchmark divides real-world dialogue into six scenarios, including a broader spectrum of contexts and more varied user intents, similar to the approach in \citet{zhu_rageval_2024}, could improve its generalizability and better challenge models. Enhancing the evaluation methodology is another important direction. Averaging scores from multiple judge models and refining judge prompts through techniques such as self-discovery \citep{zhou_self-discover_2024} could lead to more comprehensive assessments. Furthermore, examining potential biases in judge models under the Retrieval-Augmented Dialogue setting—similar to how \citet{dubois2024length} identified AlpacaEval’s preference for longer responses—would improve consistency in scoring from judge models.

\bibliography{custom}

\clearpage

\appendix

\section{Details on the Data Generation}
\label{sec:details_on_data_generation}
\begin{figure*}[htbp]
    \centering
    \includegraphics[width=0.95\textwidth]{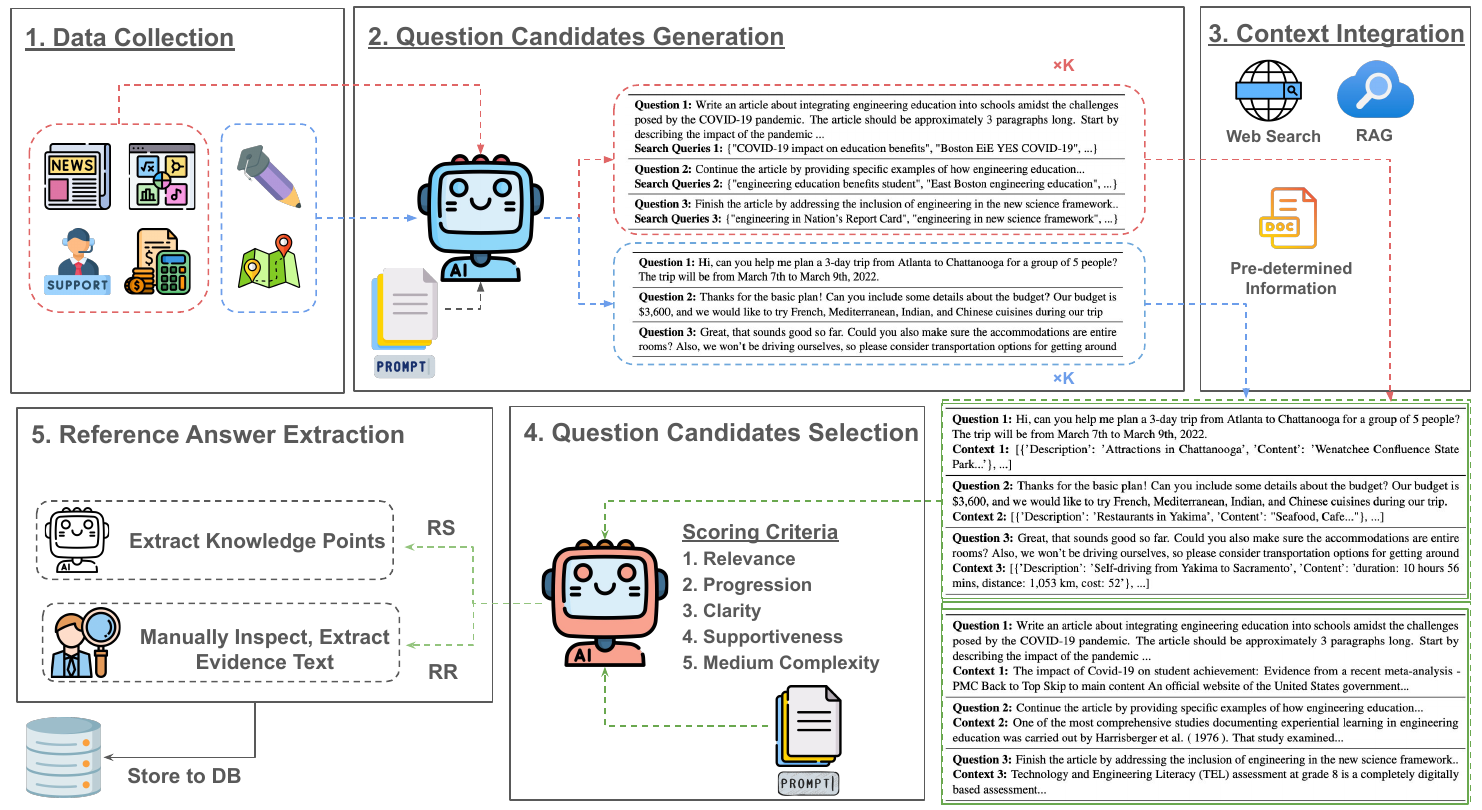}
    \caption{\textbf{Data construction pipeline of RAD-Bench}: The blue dashed lines represent scenarios with predetermined context integration at each turn, while the red dashed lines indicate scenarios where context must be retrieved via SAG or RAG, requiring additional search queries during question candidate generation (Phase 2).}
    \label{fig:rad_data_construction_pipeline}
\end{figure*}

\subsection{Data Collection}
We collect source articles and datasets from public data to form the source documents for synthetic question generation.

\textbf{Retrieval Synthesis}: For \textbf{News TLDR} scenario, we selected news articles from BBC; for \textbf{Education} scenario, we sourced popular science paragraphs from Scientific American; for \textbf{Academic Writing} scenario, we selected related work sections from papers on Arxiv and further extracted papers that appeared in each related work section. We include only source materials published after June 2024 to reduce the likelihood of the materials being included in the training data of LLMs.

\textbf{Retrieval Reasoning}: For \textbf{Customer Support} scenario, we collected user manuals from ManyManuals website. For \textbf{Finance} scenario, we leveraged datasets from FinanceBench \citep{islam_financebench_2023} as source documents. The benchmark dataset comprises 10,231 questions, answers, and evidence triplets. The evidence triplets are passages supporting answering of the question from finance report documents. We manually inspected and selected 15 triplets that involve multi-step reasoning process to get the final answer and collected corresponding source documents to serve as base data for further question candidate generation process. For \textbf{Travel Planning} scenario, we utilized TravelPlanner dataset \citep{xie_travelplanner_2024}, which comprises 1225 travel planning queries in total and leveled from simple to hard, as source documents. The hard questions in the dataset involved complicated and multiple constraints in a query, suitable for being decomposed into multi-step reasoning steps to construct instructions including constraints progressively in multi-turn dialogues. We therefore selected 15 hard questions from the training set which provides human-annotated plan as reference to serve as source data for further question candidates generation process.

\subsection{Question Candidate Generation}
\label{subsec:question_candidate_generation}
With the collected source documents, candidates of three-turn questions for each scenarios are generated by a question generator as realized by an LLM. Output of the generator for News TLDR, Education, Finance, and Customer Support scenarios for each turn includes a question and a search query. The search queries are used for retrieving relevant context as discussed in Section \ref{subsec:retrieved_context_integration}. As to Academic Writing and Travel Planning scenarios, outputs of the generator include only the questions. We craft step-by-step guidance as prompts to the generator for aligning the generated questions with the evaluated abilities. See Appendix \ref{appendix:promopts_for_qg} for details of the guidance and the prompts. We used multiple LLMs (BreeXe-8x7B, Llama3-70B, and Mixtral-8x22B) as the generator and varied the generation temperature for generating a diverse set of candidates. 

\subsection{Retrieved Context Integration}
\label{subsec:retrieved_context_integration}
In this phase, each candidate's questions for each turn are supplemented with corresponding useful information, simulating the retrieval process. For the \textbf{News TLDR} and \textbf{Education} scenarios, the accompanied search queries as produced in the question candidate generation stage are passed to the Azure web search service to retrieve the top 5 documents as useful information. For the \textbf{Customer Support} and \textbf{Finance} scenarios, we input the turn questions and source documents into Azure's RAG service to collect the retrieved contexts. For the \textbf{Academic} scenario, the information to be integrated is pre-determined. We identify referenced papers in the questions and extract the abstracts and introductions of these papers to serve as retrieved contexts for the corresponding turns. In the \textbf{Travel Planning} scenario, each turn includes reference information from the TravelPlanner bench, such as flight details, cities, and attractions, without further modification.

\subsection{Question Candidates Selection}
We employ an LLM as a scorer to assist the filtering of question candidates. For each scenario, we design customized prompts following scoring criteria to score each candidate. The criteria include Relevance, Progression, Clarity, Support, Knowledge Points, and Medium Complexity as shown in Figure \ref{fig:prompt_for_scoring_candidate}. The Support and Knowledge criteria prompt the scoring LLMs to examine whether the retrieved context from web search and RAG services contains relevant information for answering candidate questions. We scored candidates with BreeXe-8x7B, Llama3-70B, and Mixtral-8x22B. After conducting a human review of a subsampled set of scored candidates, we selected the scoring results from BreeXe-8x7B due to its preferable alignment with the established criteria. With the scored candidates of three-turn questions for each scenario, we then filtered out the top candidates and manually verified that the retrieved contexts contain informative and relevant information for answering the questions in each turn.

\subsection{Reference Answers}
\label{subsec:reference_answer}
To ensure the robustness of RAD-Bench evaluation, following the reasoning tasks in \citep{zheng_judging_2023}, we provide reference answers to benchmark questions. For evaluating scenarios in \textbf{Retrieval Synthesis}, we extract knowledge points - sets of factual statements \citep{adams_sparse_2023} - from retrieved contexts using BreeXe-8x7B as references for the first and second turn questions. As to references for the third turn, we use target paragraphs in source documents. Such reference answers thereby provide evaluator baseline quality of responses by determining whether useful knowledge points are recalled and integrated into the model's answer. For \textbf{Retrieval Reasoning}, which involves cross-turn reasoning, we manually inspect the questions and extract evidence text from the retrieved context to fully support the answers for Customer Support and Finance scenarios. In the Travel Planning scenario, we do not include reference answers in the first two turns. Instead, for the final turn, we use an expert-annotated travel plan provided in TravelPlanner Bench as the reference answer. This allows the evaluator to assess the similarity and coverage between the model's planned itinerary and the expert-annotated travel plan.

\section{Limitations}
The primary limitation of our benchmark lies in the sequential generation of questions, which may not fully capture the interdependence of dialogue turns in real-world scenarios. In the construction of RAD-Bench, benchmark questions are generated sequentially by prompting an LLM to deconstruct articles or tasks into multiple-turn questions for Retrieval Synthesis and Retrieval Reasoning, respectively. While it allows for auditable reference answers for evaluation and assesses LLMs' ability to handle changing user intents and additional constraints, it implicitly makes subsequent questions independent of earlier answers. This design lacks adaptive questioning, where users engage in ongoing dialogues due to dissatisfaction with initial LLM responses. We propose that designing follow-up questions based on the LLM's responses could create a tighter connection between rounds, better simulating real-world chatting scenarios.

Another limitation of our study is that retrieved contexts are pre-specified. While this design choice enables us to focus on the generation end to effectively evaluate how models utilize given context to handle changing user intents and additional requirements, it represents a constrained scenario within the broader retrieval-augmented dialogue (RAD) pipeline encountered in real-world applications. future research aimed at benchmarking the entire end-to-end RAD pipeline may provide insights into potential areas for comprehensive system improvements.

\section{Performance of evaluated LLMs}
\begin{figure}[!htbp]
  \includegraphics[width=\columnwidth]{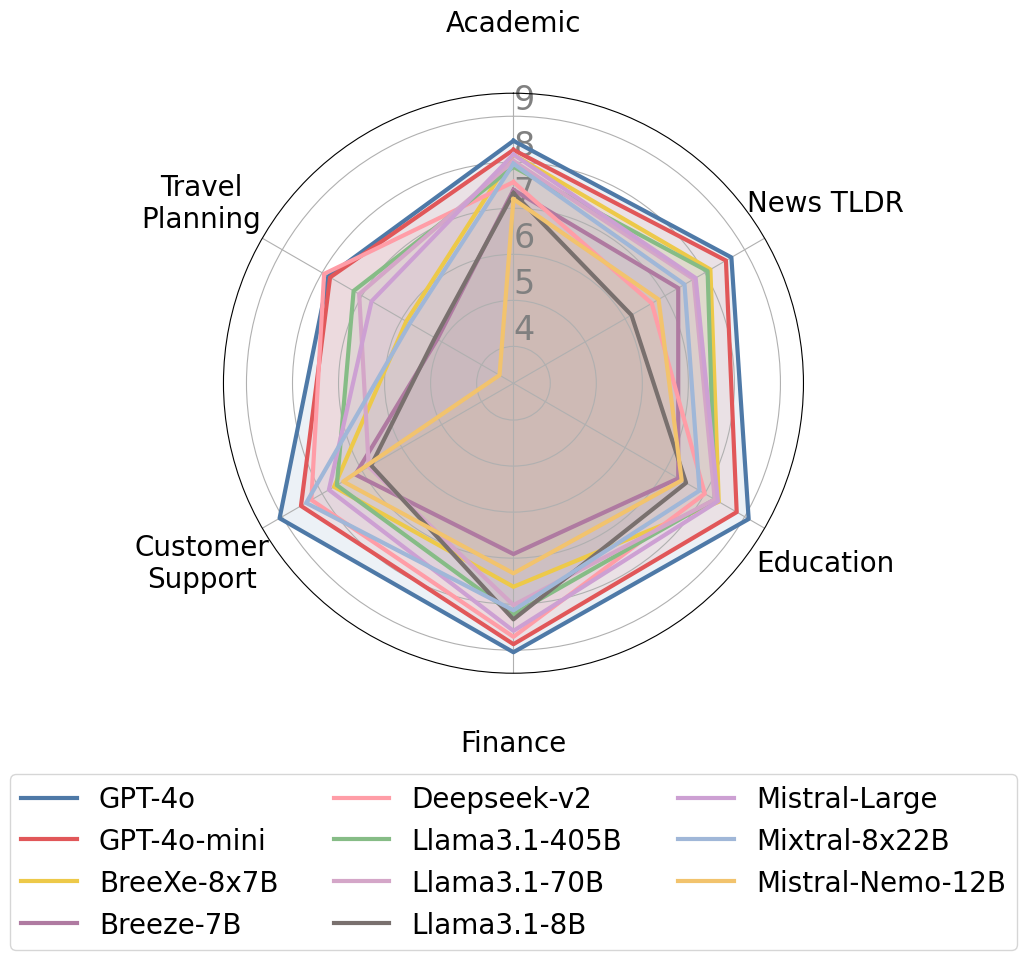}
  \caption{\textbf{Performance of evaluated LLMs}}
  \label{fig:model_result_radar}
\end{figure}
\vspace{-5pt}

\newpage
\section{Evaluated aspects and selected application scenarios}
\begin{figure}[htbp]
    \centering
    \includegraphics[width=0.7\columnwidth]{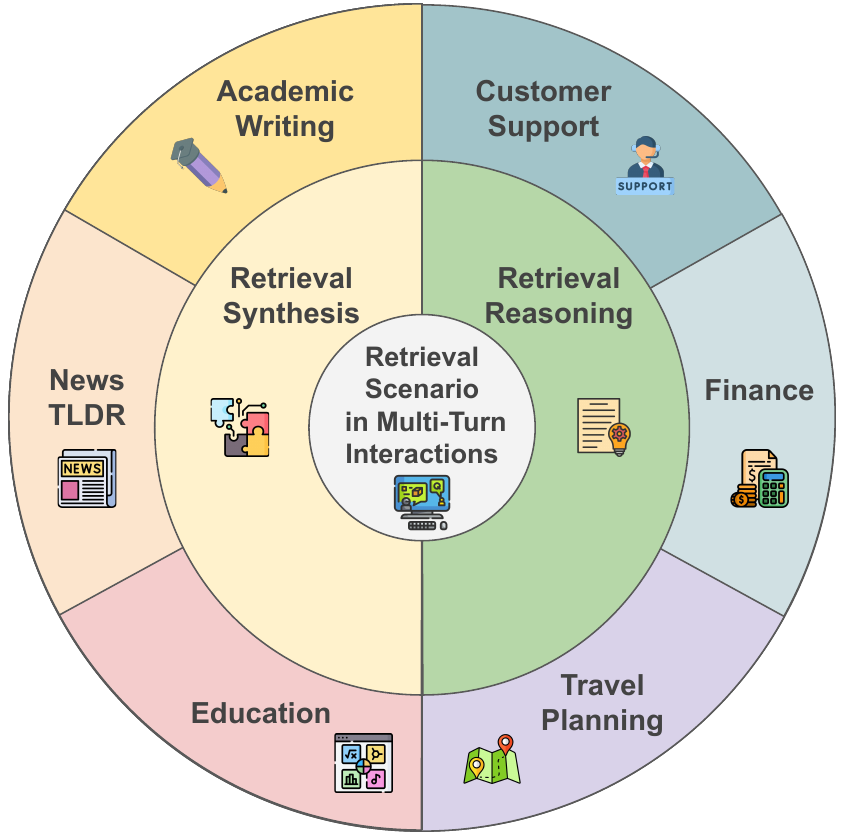}
    \caption{Evaluated capabilities—\emph{Retrieval Synthesis} and \emph{Retrieval Reasoning}—across three concrete application scenarios each. See Appendix \ref{appendix:dialogue_examples} for examples of augmented dialogues following retrievals.}
    \label{fig:rad_hierarchy}
\end{figure}
\vspace{-5pt}

\onecolumn
\section{Performance of models across model sizes}

\begin{figure*}[htbp]
    \centering
    \includegraphics[width=\textwidth]{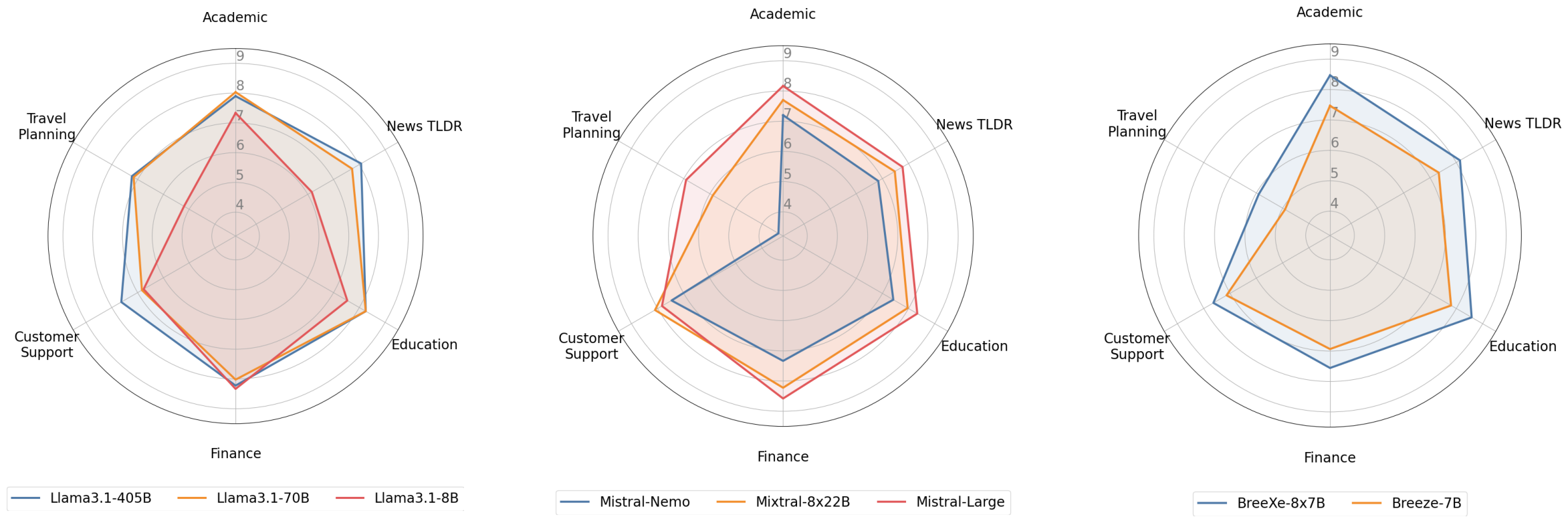}
    \caption{Performance of various LLMs by categories (Llama 3.1, Mistral, and Breeze/BreeXe)}
    \label{fig:combined_radar_charts}
\end{figure*}

\newpage
\onecolumn

\section{Prompts for Question Generation}
\label{appendix:promopts_for_qg}

\begin{minipage}{\textwidth}
    \centering
    \setlength{\fboxsep}{5pt}
    
    \fbox{
        \parbox{0.9\textwidth}{
            \textbf{system\_prompt:} You are an experienced writer tasked with designing a series of connected queries to guide an AI in progressively summarizing, comparing, and analyzing key points of an event or story.
            The goal is to integrate new context at each step, resulting in a comprehensive summarization (TL;DR, tables, bullet points, analysis, etc.) that can cover as many key points as possible from a source article. To complete this, follow the following instructions:
            
            [The Start of Instruction]\\
            1. Identify key knowledge points in the source article that are crucial for understanding the event or story.\\
            2. Design the first turn query:\
            - Decide on the final output format (e.g., TL;DR, comparison table, bullet points).\\
            - Specify the desired length and structure of the output (e.g., word count, number of paragraphs).\\
            3. Design the second and third turn query:\\ - Identify additional context or background information that can enhance the initial draft.\\
            - Guide the AI to integrate this new information into the existing draft.\\
            Guide the AI to incorporate this analysis into the current draft.\\
            - Include relevant web search queries to gather expert opinions and analysis
            
            [The End of Instruction]
            
            Below are some important requirements you need to strictly follow when generating the three-turn question set:
            
            [The Start of Important Requirements]\\
            1. In the first turn, the query needs to guide the AI to specify what the final output should look like. (e.g., writing comparison table, writing TL;DR, bullet points, ...)\\
            2. In the second and third turn, do not specify the output format\\
            3. Emphasize the continuity of the questions, prompting the AI to keep working on the current draft and adding knowledge points progressively.\\
            4. Avoid asking the AI to generate a whole new article in each turn\\
            5. Ensure the tasks are diverse, such as generating a comparison table, creating bullet points, and writing a brief analysis, rather than just writing a TL;DR.\\
            6. Please Strictly follow the specified output JSON format (in the end of the instruction) for the three-turn question set you come up with.
            
            [The End of Important Requirements]
            
            For the design of a set with connected questions and relevant web search queries, you can refer to the following example:
            
            [The Start of Examples]
            \{few\_shot\_learning\_text\}
            [The End of Examples]
 
            \textbf{prompt\_template:} The following is the article you need to carefully read and generate questions for: [The Start of The Article]
            \{source\_doc\}
            [The End of The Article]\\
            Remember in the first turn's query, you should specify what needs to be done by the AI (the final output, e.g., TL;DR summary, comparison table, bullet points, etc.). YOU CANNOT DESIGN QUESTIONS THAT ARE SIMILAR TO QUESTIONS GENERATED IN PREVIOUS ROUNDS.
            As for the final question set output format, YOU SHOULD STRICTLY FOLLOW THE FOLLOWING OUTPUT JSON FORMAT:
            
            [The Start of the OUTPUT JSON FORMAT]
            \{output\_format\}
            
            [The End of the OUTPUT JSON FORMAT]\\
            You need to STRICTLY FOLLOW the specified output JSON format to serve as your FINAL OUTPUT!

            \textbf {output\_format:}
            [\{"query": "....", "answer": "...", "referenced\_information": "..."\}, \{"query": "....", "answer": "...", "referenced\_information": "..."\}, \{"query": "....", "answer": "...", "referenced\_information": "..."\}]
        }

    }
    \captionof{figure}{The prompt to generate questions of News TLDR scenario.}
\end{minipage}
\newpage

\begin{minipage}{\textwidth}
    \centering
    \setlength{\fboxsep}{5pt}
    
    \fbox{
        \parbox{0.9\textwidth}{
            \textbf{system\_prompt:} "You are an experienced writer tasked with designing a series of connected queries to guide an AI in progressively generating a draft article. The goal is to integrate new context at each step, resulting in a comprehensive final article. Each query should focus on one main aspect, ensuring the AI can build upon the previous draft with new information. Include relevant web search queries to help gather necessary information for each turn. To achieve this, follow these steps:\\
            1. Identify several main knowledge points in the provided article.\\
            2. Group the knowledge points into three main aspects .\\
            3. Design each query to focus on one aspect at a time, ensuring that the AI can integrate new information progressively.\\
            4. Ensure each query builds upon the previous draft, adding layers of information from different references.\\
            5. Include a list of relevant web search query, each focuses on designing a web search query that can gather necessary information the turn needs for answering correctly. The search query list should have exactly 3 queries. Output the 3 connected queries in JSON format, where each query entry should include:\\
            1. "query": The query for the AI to generate the draft article.\\
            2. "web\_search\_query": A list of highly relevant web search query to find articles that can help construct the specified draft article. What needs to be noticed is that the query should only focus on one aspect at a time, and DO NOT ask questions that involves multiple actions such as summarize and compare at the same time.
            
            [Important Requirements]\\
            1. In the first turn, the first turn's query needs to guide the AI to specify what the final output should look like (e.g., word count, paragraph count, what needs to be done, etc.) and include the instruction to follow the specified output format. For example, the first turn's query can start with: "I want to write an article about ... The draft should be around ... paragraphs, ... words, etc."\\
            2. In the second and third turn, do not specify the output format!.\\
            3. Emphasize the continuity of the questions, prompting the AI to keep working on the current draft and adding knowledge points progressively.\\
            4. Avoid asking the AI to generate a whole new article in each turn.\\
            For the design of a set with connected questions and relevant web search queries, you can refer to the following example: [The Start of Examples] \{few\_shot\_learning\_text\} [The End of Examples]\\
 
            \textbf{prompt\_template:} The following is the article you need to carefully read and generate questions for: [The Start of The Article]\{source\_doc\}
            [The End of The Article]\\
            You should strictly follow the following output JSON format: {output\_format}.
            
            \textbf {output\_format:}
            [\{"query": "....", "answer": "...", "referenced\_information": "..."\}, \{"query": "....", "answer": "...", "referenced\_information": "..."\}, \{"query": "....", "answer": "...", "referenced\_information": "..."\}]
        }

    .}
    \captionof{figure}{The prompt to generate questions of Education scenario.}
\end{minipage}
\newpage

\begin{minipage}{\textwidth}
    \centering
    \setlength{\fboxsep}{5pt}

    \fbox{
        \parbox{0.9\textwidth}{
            \textbf{system\_prompt:} "You are an experienced academic writer with expertise in constructing "Related Work" sections for research papers. Now given a related work's paragraph, what you need to do is to design a series of three connected queries that will guide an AI to reconstruct the related work section progressively, integrating new context at each step to build a comprehensive final draft. In this task, you need to focus on identifying several key information points, grouping them into three main aspects, and ensuring that each query explicitly prompts the AI to expand upon a working draft "Related Work" section based on new information gathered at each step. Each query should guide the AI to build further on the previous draft, connecting the three main aspects. Additionally, for each question, identify those references that can be used to support the content by providing a list of reference\_id.\\
            
            To achieve this, follow these steps:\\
            1. Identify several key information points in the provided related work section.\\
            2. Group the key information points into three main aspects.\\
            3. Design each query to focus on one aspect at a time, ensuring that the AI can integrate new information progressively.\\
            4. Ensure each query builds upon the previous draft, adding layers of information from different references.\\
            5. Include a list of relevant reference\_ids for each query, ensuring that the references are used to support the content and are not empty. \\
            Output the 3 connected queries in JSON format, where each query entry should include:\\
            1. "query": The query for the AI to generate the draft "Related Work" section.\\
            2. "reference\_ids": A list of reference IDs that are mentioned in the query and can be used to support the question. \\
            Please make sure you directly output the JSON format but not one query at a time.\\
 
            \textbf{prompt\_template:} As an experienced academic writer specializing in education and related fields, you are tasked with designing three connected queries that will guide an AI to progressively generate a draft "Related Work" section for a research paper. Each query should build upon the previous one by integrating new context and insights, ultimately creating a comprehensive and cohesive final draft. The following article is provided as a source document for you to carefully review and design the questions:
            \{source\_doc\}\\
            YOU CANNOT DESIGN QUESTIONS THAT ARE SIMILAR TO QUESTIONS GENERATED IN PREVIOUS ROUNDS. YOU SHOULD STRICTLY FOLLOW THE FOLLOWING OUTPUT JSON FORMAT:
            \{output\_format\}\\
            The above output is just for your reference, you really need to carefully generate the query and corresponding reference ids list for the query ensuring these ids are all valid and existed in the given related work section. Please make sure you directly output the JSON format but not one query at a time.\\
            
            \textbf {output\_format:}
            [\{"query": "....", "answer": "...", "referenced\_information": "..."\}, \{"query": "....", "answer": "...", "referenced\_information": "..."\}, \{"query": "....", "answer": "...", "referenced\_information": "..."\}]
        }

    }
    \captionof{figure}{The prompt to generate questions of Academic scenario.}
\end{minipage}
\newpage

\begin{minipage}{\textwidth}
    \centering
    \setlength{\fboxsep}{5pt}
    
    \fbox{
        \parbox{0.9\textwidth}{
            \textbf{system\_prompt:} You are a helpful and logical assistant specialized in finance and data analysis. Your task is to help users break down complex finance-related questions into simpler, intermediate questions that logically lead to a final question. Ensure that the answers provided are accurate and based on the given evidence text. You will be provided with information texts, and you need to generate a sequence of three questions and answers that build up to the final correct question and answer with the appropriate evidence text. For the design of the three connected follow-up questions, you can refer to the following examples:
            \{few\_shot\_learning\_text\}.
            \\
            
            \textbf{prompt\_template:}
            Given the following expert-designed finance question, answer, and evidence text, think step by step and generate three questions with their answers and evidence text that can be built to lead to the final correct question and correct answer with the correct evidence text.
            [The Start of the Given Document] \# source\_doc \# [The End of the Given Document]
            \vspace{0.3cm}
            
            You need to follow the below instructions to construct the data:
            
            [The Start of Instruction ] \\
            1. Identify Key Components: Break down the main question into its key components (e.g., time periods, specific events, financial metrics).\\
            2. Logical Steps: Determine the logical steps required to answer the main question. Each step should build on the previous one and lead to the final question.\\
            3. Generate Intermediate Questions: Create intermediate questions that address each logical step. Ensure each question is neither too easy nor too difficult and that it logically connects to the next question.\\
            4. Reference Evidence Text: Ensure each question can be answered using the provided evidence text. Clearly reference the part of the text that supports the answer. It has to be clear and you need to really make sure the question you propose can be answered or inferred from the support text you extracted\\
            5. Final Question: Use the answers from the intermediate questions to generate the final question, ensuring it matches the provided final question and answer. The final question should be the same or very similar to the provided main question to ensure it is the most difficult part
            
            [The End of Instruction]\\
            You should strictly follow the following output JSON format: \{output\_format\}.\\
            
            \textbf{output\_format:}
            [\{"query": "....", "answer": "...", "referenced\_information": "..."\}, \{"query": "....", "answer": "...", "referenced\_information": "..."\}, \{"query": "....", "answer": "...", "referenced\_information": "..."\}]
        }
    }
    \captionof{figure}{The prompt to generate questions of Finance scenario.}
\end{minipage}
\newpage

\begin{minipage}{\textwidth}
    \centering
    \setlength{\fboxsep}{5pt}
    
    \fbox{
        \parbox{0.9\textwidth}{
            \textbf{system\_prompt:} 
            You are an experienced customer support agent who can handle user queries effectively by progressively narrowing down the problem and using reasoning techniques to identify the root cause. You will be provided with a user manual containing common errors and solution suggestions. Your task is to design three connected dialogue turns that simulate a user talking to a customer support agent to solve problems they encounter. Each turn should include a user question, context that supports answering the question, and a precise agent answer. The questions should progressively scope down and test the agent's ability to reason and figure out the root cause of the user's problem. The initial query might be broad and vague, the second turn should follow the agent's solution but still encounter some problems, and the final turn should further narrow down the possible cause by providing new evidence. The final turn should correctly identify the problem the user encounters. To achieve this, follow these steps:\\
            1. Identify a common error from the user manual and its suggested solutions.\\
            2. Create a broad initial user query based on the common error.\\
            3. Design the second user query to follow up on the agent's initial response, indicating that the initial solution did not fully resolve the issue and providing additional details or symptoms.\\
            4. Design the third user query to provide new findings or evidence based on the previous troubleshooting steps, leading to a more specific troubleshooting step or final resolution.\\
            5. Ensure each agent answer is clear, precise, and directly addresses the user's issue.\\
            6. Extract the context directly from the user manual to support each answer.\\
            Output the three connected dialogue turns in JSON format, where each entry should include:\\
            1. "query": The user's question.\\
            2. "context": The extracted context from the user manual that supports answering the question.\\
            3. "answer": The agent's response.\\

            \textbf{prompt\_template:}
            Here is the provided user manual: 
            [The Start of Manual] \{source\_doc\} [The End of Manual]. Read it carefully and try to identify a common error and its suggested solutions. Based on this, design three connected dialogue turns that simulate a user talking to a customer support agent to solve the problem they encounter. Each turn should include a user question, context that supports answering the question, and a precise and clear agent answer. The questions should progressively scope down and test the agent's ability to reason and figure out the root cause of the user's problem. The initial query might be broad and vague, the second turn should follow the agent's solution but still encounter some problems, and the final turn should further narrow down the possible cause by providing new findings or evidence. The final turn should correctly identify the problem the user encounters. Output the three connected dialogue turns in JSON format, where each entry should include:\\
            1. "query": The user's question.\\
            2. "context": The context that supports answering the question SHOULD BE DIRECTLY EXTRACTED FROM THE USER MANUAL, WHICH IS A PIECE OF INFORMATION IN THE MANUAL. YOU NEED TO MAKE SURE THE CONTEXT IS HELPFUL FOR ANSWERING THE QUESTIONS\\
            3. "answer": The agent's response.\\
            REMEMBER: YOU CANNOT DESIGN QUESTIONS THAT ARE SIMILAR TO QUESTIONS GENERATED IN PREVIOUS ROUNDS. IT MEANS THAT YOU HAVE TO IDENTIFY NEW PROBLEMS AND TRY TO USE THAT FOR CONSTRUCTING THE THREE TURN QUESTION SET. IN THE END, YOU SHOULD STRICTLY FOLLOW THE FOLLOWING OUTPUT JSON FORMAT:
            \{output\_format\}\\
            Please Make sure you really directly output the JSON format but not one query at a time!\\
            
            \textbf{output\_format:} [{"query": "....", "answer": "...", "context": "..."}, {"query": "....", "answer": "...", "context": "..."}, {"query": "....", "answer": "...", "context": "..."}]
        }
    }
    \captionof{figure}{The prompt to generate questions of Customer Support scenario.}
\end{minipage}
\newpage

\begin{minipage}{\textwidth}
    \centering
    \setlength{\fboxsep}{5pt}
    
    \fbox{
        \parbox{0.9\textwidth}{
            \textbf{system\_prompt:} You are a helpful and logical assistant specialized in travel planning. Your task is to help users break down complex travel-related queries into simpler, intermediate queries that logically lead to a final, more complex query. Ensure that the plans provided are accurate and based on the given reference information. You will be provided with information texts, and you need to generate a sequence of three queries that build up to the final correct query with the appropriate reference information.\\
            
            \textbf{prompt\_template:} You are a helpful and logical assistant specialized in travel planning. Your task is to help users break down complex travel-related queries into simpler, intermediate queries that logically lead to a final, more complex query. Ensure that the plans provided are accurate and based on the given reference information. You will be provided with information texts, and you need to generate a sequence of three queries that build up to the final correct query with the appropriate reference information. You will be given the original complex query and corresponding annotated constraints. What you need to do is to generate a three-turn question set starting from basic requirements, progressively adding constraints to build up to the final turn containing all constraints. Each query should build on the previous one without repeating the requirements already mentioned. Each query should prompt the AI to generate a complete plan based on the given constraints. The queries should be natural and conversational, just like a user talking to a travel agent. You have to strictly follow the output format: \{output\_format\}\\
            
            \textbf{output\_format:}
            [{"query": "....", "constraints": {}},{"query": "....", "constraints": {}},{"query": "....", "constraints": {}}
            ]
        }
    }
    \captionof{figure}{The prompt to generate questions of Travel Planning scenario.}
\end{minipage}
\newpage

\begin{minipage}{\textwidth}
    \centering
    \setlength{\fboxsep}{5pt}
    
    \fbox{
        \parbox{0.9\textwidth}{
            Please act as an impartial judge and evaluate the quality of the generated three-turn question set based on the source document provided. Your evaluation should consider factors such as relevance, progression, clarity, support, and knowledge points. The explanation of these factors are given below:
            \vspace{0.3cm}
            
            - Relevance: How closely the questions align with the source document and the task prompt\\
            - Progression: How well each question builds upon the previous one to add new layers of information.\\
            - Clarity: The clarity and unambiguity of the questions\\
            - Support: The relevance and utility of the suggested web search queries or reference IDs\\
            - Knowledge Points: How well the key information retrieved from the specified web search queries can be utilized in the questions.\\
            - Medium Complexity: The question needs to be focused and do not involve too many perspectives in one time!! Simply to say, a good question should focus on certain aspects but never cover too many knowledge points. That is to say, if a question covers too many topics, aspects at a time, you should see this as a question that is too difficult and deduct some points.
            \vspace{0.3cm}
            
            Now carefully review the source document provided and the answer generated:
            
            [The Start of Original Article]
            \{reference\} 
            [The End of Original Article]
            \vspace{0.3cm}
            
            [The Start of Three-Turn Question Set to be evaluated]: \{answer\} [The End of Three-Turn Question Set to be evaluated]
            \vspace{0.3cm}
            
            Begin your evaluation by providing a short explanation. Be as objective as possible. After providing your careful and comprehensive explanation, you must rate the question set on a scale of 1 to 5 by strictly following this format: "<FINAL>[[rating]]</FINAL>", for example: "Rating: <FINAL>[[4]]</FINAL>"
        }
    }
    \captionof{figure}{The prompt for the scoring candidates.}
    \label{fig:prompt_for_scoring_candidate}
\end{minipage}

\newpage
\section{Prompts for Evaluation}
\label{appendix:promopts_for_eval}
\begin{minipage}{\textwidth}
    \centering
    \setlength{\fboxsep}{5pt}
    
    \fbox{
        \parbox{0.9\textwidth}{
            [Instruction]\\
            Please act as an impartial judge and evaluate the quality of the response provided by an AI assistant to the user question displayed below. \\
            Your evaluation should consider helpfulness and Informativeness:\\
            
            [Helpfulness] \\
            you should evaluate the helpfulness of the assistant's answer to the question of current turn.
        
            [Informativeness] \\
            You are given the assistant's answer and reference knowledge points representing knowledge that should be mentioned, discussed, and covered in the assistant's answer. You should evaluate how informativeness the assistant's answer is in including the reference knowledge points appropriately. \\
            Begin your evaluation by comparing the assistant's answer with the reference knowledge points. Be as objective as possible. After providing your explanation, you must rate the response on a scale of 1 to 10 by strictly following this format: "[[rating]]", for example: "Rating: [[5]]".\\
            
            [Question]\\
            \{question\}
            
            [End of Question]\\
            
            [The Start of Reference Knowledge Points]\\ 
            \{reference\}
            
            [The End of Reference Knowledge Points]\\
            
            [The Start of Assistant's Answer]\\
            \{answer\}
            
            [The End of Assistant's Answer]
        }
    }
    \captionof{figure}{Prompt for evaluating the first turn of a scenario in Retrieval Synthesis.} 
\end{minipage}

\newpage
\newpage
\begin{minipage}{\textwidth}
    \centering
    \setlength{\fboxsep}{5pt}
    
    \fbox{
        \parbox{0.9\textwidth}{
            [Instruction]\\
            Please act as an impartial judge and evaluate the quality of the response provided by an AI assistant to the user question displayed below. 
            
            Your evaluation should assess the helpfulness, coherence, adherence, and informativeness:\\
            
            [Helpfulness] you should evaluate the helpfulness of the assistant's answer to the question of current turn. \\
            
            [Informativeness] You are given the assistant's answer and reference knowledge points representing knowledge that should be mentioned, discussed, and covered in the assistant's answer. You should evaluate how informativeness the assistant's answer is in including the reference knowledge points appropriately. \\
            
            [Adherence] You are given question of the previous turn. Consider how well the assistant's answer respects the user intents throughout the turns.\\
            
            [Coherence] you are given the user questions and reference knowledge points in the previous turns to serve as previous instructions. You should consider how well the assistant's answer aligns with the knowledge points mentioned in the current turn's reference knowledge points and how it respects or builds upon the focus and knowledge points from the previous turns. \\
            
            Begin your evaluation by comparing the assistant's answer against the reference knowledge points from both previous and current turns. Be as objective as possible, and provide a detailed justification for your rating. After providing your explanation, you must rate the response on a scale of 1 to 10, strictly following this format: "Rating: [[rating]]", for example: "Rating: [[5]]".\\
            
            [The Start of Previous Questions and Reference Knowledge Points]\\
            Question: \{question\_1\}\\
            Reference Knowledge Points: \{reference\_1\}
            
            [The End of Previous Questions and Reference Knowledge Points]\\
            
            [The Start of Current Turn Question]\\
            \{question\}
            
            [The End of Current Turn Question]\\
            
            [The Start of Reference Knowledge Points]\\
            \{reference\}
            
            [The End of Reference Knowledge Points]\\
            
            [The Start of Assistant's Answer]\\
            \{answer\}
            
            [The End of Assistant's Answer]
        }
    }
    \captionof{figure}{Prompt for evaluating the second turn of a scenario in Retrieval Synthesis.} 
\end{minipage}

\newpage

\begin{minipage}{\textwidth}
    \centering
    \setlength{\fboxsep}{5pt}
    
    \fbox{
        \parbox{0.9\textwidth}{
            [Instruction]\\
            Please act as an impartial judge and evaluate the quality of the response provided by an AI assistant to the user question displayed below. Your evaluation should assess the correctness, helpfulness. Your evaluation should focus on the assistant's answer to the question of current turn. You also need to evaluate the adherence of the assistant's answer to previous instructions. You will be given the assistant's answer and a reference answer. You will also be given the user questions and reference knowledge points in the previous turns to serve as previous instructions. You should consider how well the assistant's answer captures the key information, knowledge points mentioned in the reference answer and how it respects or builds upon the focus and knowledge points from the previous turns. 
            \\
            
            Your evaluation should assess the helpfulness, coherence, adherence, and informativeness:
            \\
            
            [Helpfulness] \\
            you should evaluate the helpfulness of the assistant's answer to the question of current turn. 
            \\
            
            [Informativeness] \\
            You are given the assistant's answer and reference knowledge points representing knowledge that should be mentioned, discussed, and covered in the assistant's answer. You should evaluate how informativeness the assistant's answer is in including the reference knowledge points appropriately.
            \\
            
            [Adherence]\\
            You are given questions of the previous turns. Consider how well the assistant's answer respects the user intents throughout the turns.
            \\
            
            [Coherence]\\
            you are given the user questions and reference knowledge points in the previous turns to serve as previous instructions. You should consider how well the assistant's answer aligns with the knowledge points mentioned in the current turn's reference knowledge points and how it respects or builds upon the focus and knowledge points from the previous turns.
            \\

            Begin your evaluation by comparing the assistant's answer against the reference answer in this turn and reference knowledge points in previous turns. Be as objective as possible, and provide a detailed justification for your rating. After providing your explanation, you must rate the response on a scale of 1 to 10, strictly following this format: "Rating: [[rating]]," for example: "Rating: [[5]]".
        }
    }
    \captionof{figure}{Prompt for evaluating the final turn of a scenario in Retrieval Synthesis.} 
\end{minipage}
\vspace{0.5cm}

\newpage
\begin{minipage}{\textwidth}
    \centering
    \setlength{\fboxsep}{5pt}
    
    \fbox{
        \parbox{0.9\textwidth}{
            [Instruction]\\
            Please act as an impartial judge and evaluate the quality of the response provided by an AI assistant to the user question displayed below. Your evaluation should consider correctness, helpfulness, and reasoning correctness. Additionally, you need to assess how effectively the assistant utilizes the given context to generate its response. The assistant’s answer should align with the provided context and avoid any factual inaccuracies or hallucinations that cannot be inferred from the given context. You will be given a reference answer representing a correct response, context the assistant needs to utilize and the assistant's answer. Begin your evaluation by comparing the assistant's answer with the reference answer and considering its adherence to the context.\\
            Be as objective as possible. After providing your explanation, you must rate the response on a scale of 1 to 10 by strictly following this format: "Rating: [[rating]]", for example: "Rating: [[5]]".\\
            
            [Question]\\
            \{question\}
            
            [The Start of Context]\\
            \{context\}
            
            [The End of Context]\\
            
            [The Start of Reference Answer]\\
            \{reference\}
            
            [The End of Reference Answer]\\
            
            [The Start of Assistant's Answer]\\
            \{answer\}
            
            [The End of Assistant's Answer]
        }
    }
    \captionof{figure}{Prompt for evaluating the first turn of a scenario in Retrieval Reasoning.} 
\end{minipage}

\newpage
\begin{minipage}{\textwidth}
    \centering
    \setlength{\fboxsep}{5pt}
    
    \fbox{
        \parbox{0.9\textwidth}{
            Please act as an impartial judge and evaluate the quality of the response provided by an AI assistant to the question of current turn displayed below. Your evaluation should consider correctness, helpfulness, and reasoning correctness. Additionally, assess how effectively the assistant utilizes the given context and adheres to constraints from both the first and the current turn to generate its response. The assistant's answer should align with the provided context from current turn and avoid any factual inaccuracies or hallucinations that cannot be inferred from the given context. You will be given a conversation history in previous turns to evaluate the adherence of the assistant's answer in the current turn. You will also be given a reference answer representing a correct response, context the assistant needs to utilize and the assistant's answer. Begin your evaluation by comparing the assistant's answer with the reference answers from both turns and considering its adherence to the context and logical progression.
            \\
            
            Be as objective as possible. After providing your explanation, you must rate the response on a scale of 1 to 10 by strictly following this format: "Rating: [[rating]]", for example: "Rating: [[5]]".
            \\
            
            [The Start of Original Article]\\
            \{reference\}
            
            [The End of Original Article]
            \\

            [The Start of The Conversation History]\\
            User: \{question\_1\}\\
            Assistant's Answer: \{reference\_1\}\\
            User: \{question\_2\}\\
            Assistant's Answer: \{reference\_2\}
            
            [The End of The Conversation History]
            \\

            [The Start of Current Turn Question]\\
            \{question\}
            
            [The End of Current Turn Question]
            \\
            
            [The Start of Current Turn Context]\\
            \{context\}
            
            [The End of Current Turn Context]
            \\

            [The Start of Current Turn Reference Answer]\\
            \{reference\}
            
            [The End of Current Turn Reference Answer]
            \\
            
            [The Start of Assistant's Answer]\\
            \{answer\}
            
            [The End of Assistant's Answer]
        }
    }
    \captionof{figure}{Prompt for evaluating the second turn of a scenario in Retrieval Reasoning.} 
\end{minipage}

\newpage
\begin{minipage}{\textwidth}
    \centering
    \setlength{\fboxsep}{5pt}
    
    \fbox{
        \parbox{0.9\textwidth}{
            Please act as an impartial judge and evaluate the quality of the response provided by an AI assistant to the question of current turn displayed below. Your evaluation should consider correctness, helpfulness, and reasoning correctness. Additionally, assess how effectively the assistant utilizes the given context and adheres to constraints from both the first and the current turn to generate its response. The assistant's answer should align with the provided context from current turn and avoid any factual inaccuracies or hallucinations that cannot be inferred from the given context. You will be given a conversation history in previous turns to evaluate the adherence of the assistant's answer in the current turn. You will also be given a reference answer representing a correct response, context the assistant needs to utilize and the assistant's answer. Begin your evaluation by comparing the assistant's answer with the reference answers from both turns and considering its adherence to the context and logical progression.
            \\
            
            Be as objective as possible. After providing your explanation, you must rate the response on a scale of 1 to 10 by strictly following this format: "Rating: [[rating]]", for example: "Rating: [[5]]".
            \\
            
            [The Start of Original Article]\\
            \{reference\}
            
            [The End of Original Article]
            \\

            [The Start of The Conversation History]\\
            User: \{question\_1\}\\
            Assistant's Answer: \{reference\_1\}\\
            User: \{question\_2\}\\
            Assistant's Answer: \{reference\_2\}
            
            [The End of The Conversation History]
            \\

            [The Start of Current Turn Question]\\
            \{question\}
            
            [The End of Current Turn Question]
            \\
            
            [The Start of Current Turn Context]\\
            \{context\}
            
            [The End of Current Turn Context]
            \\

            [The Start of Current Turn Reference Answer]\\
            \{reference\}
            
            [The End of Current Turn Reference Answer]
            \\
            
            [The Start of Assistant's Answer]\\
            \{answer\}
            
            [The End of Assistant's Answer]
        }
    }
    \captionof{figure}{Prompt for evaluating the final turn of a scenario in Retrieval Reasoning.} 
\end{minipage}

\newpage
\section{Multi-turn dialogues examples  for different scenarios}
\label{appendix:dialogue_examples}
\begin{figure*}[h!]
    \centering
    \groupedRowColors[0]{3}{1}{orange!20}{white}
    
    \begin{tabular}{ p{15cm} }
        \toprule
        \textbf{Question 1:} Help me write a related work section. Begin by discussing the trend of porting various data structures to become GPU-resident in recent years. Explain the general approach of optimizing memory layout and access patterns to fit the underlying GPU architecture. Mention some examples of GPU-resident indexes, such as hash tables, bloom filters, and quotient filters, and their typical use cases. \\
        \textbf {Context 1:} GPU hash tables are useful for interactive graphics applications... Properties such as an amortized constant time complexity for insertion and querying as well as a compact memory layout make them versatile associative data structures with manifold applications... More recent approaches including cuDF [8]... \\
        \midrule
        \textbf{Assistant:} In recent years, there has been a growing interest in adapting and optimizing various data structures to take advantage of the massive parallelism and fast memory interfaces available on modern GPUs... 
        Hash tables, in particular, have been widely used in various applications...
        Several approaches have been proposed to accelerate hash map construction on GPUs...
        In summary, the trend of porting various data structures to become GPU-resident has been driven by the need for accelerated data processing pipelines and the optimization of memory layout and access patterns to fit the underlying GPU architecture...\\
        \midrule
        \textbf{Question 2:} Next, expand the draft and focus on GPU-resident tree structures, such as radix trees and comparison-based trees. Highlight their capabilities, including range-lookup support. Mention that while the evaluation includes a state-of-the-art comparison-based tree B+, no code for the radix tree is available. \\
        \textbf{Context 2:} Application-specific data structures, such as acceleration tree data structures used in ray tracing [31], may also be useful for general-purpose tasks... Furthermore, B-Tree in- sertions are also faster than LSM and sorted array insertions...Particularly challenging is the development of dynamic (mu- table) data structures that can be built, queried, and updated on the GPU... \\
        \midrule
        \textbf{Assistant:} In the realm of GPU-resident tree structures, two main categories can be distinguished: radix trees and comparison-based trees...
        Radix trees, also known as prefix trees or Patricia tries, are a versatile data structure used for storing a set of strings, providing efficient lookups and range queries...
        Comparison-based trees, such as B-trees, red-black trees, and B+ trees, are another type of GPU-resident tree structure...
        It is important to note that while our evaluation includes a state-of-the-art comparison-based tree B+, no code for the radix tree is available...
        \\
        \midrule
        \textbf{Question 3:} Finish the draft by discussing GPU-resident spatial indexes, such as R-Trees, and their potential benefits for constructing bounding volumes. Note the absence of available code for these indexes, which limits their use as baselines for comparisons. \\
        \textbf{Context 3:} An R-tree is a data structure for organizing and querying multi-dimensional non-uniform and overlapping data... R-Trees are popular spatial indexing techniques that have been widely adopted in many geospatial applications... \\
        \midrule  
        \textbf{Assistant:} In conclusion, the development of GPU-resident spatial indexes, such as R-Trees, holds great potential for accelerating various computational tasks that involve the processing of multi-dimensional non-uniform and overlapping data...
        R-Trees, in particular, have been demonstrated to deliver significant performance improvements when implemented on GPUs...
        However, it is important to note that the absence of available code for these GPU-resident indexes, such as R-Trees, limits their use as baselines for comparisons...
        As society continues to rely on geospatial applications and the demand for high-performance, GPU-accelerated applications grows, the development of efficient and adaptable GPU-resident spatial indexes is of paramount importance...\\
        \midrule
    \end{tabular}
    \caption{Multi-turn dialogues example for Academic scenario}
    \label{fig:rad_example_academic}
\end{figure*}

\newpage
\begin{figure*}[h!]
    \centering
    \groupedRowColors[0]{3}{1}{orange!20}{white}
    
    \begin{tabular}{ p{15cm} }
        \toprule
        \textbf{Question 1:} Write a detailed article about the Chicago Teachers Union's approach towards climate change in their contract negotiations. The article should be approximately 3-4 paragraphs long. Start by explaining the background of the CTU and their previous successes in negotiating contracts. Then, describe Stacy Davis Gates' strategy of 'bargaining for the common good,' including climate policies like installing solar panels and heat pumps in schools. Finally, touch upon the opposition's viewpoints and the potential impact of the CTU's efforts on the city and state budget. \\
        \textbf {Context 1:} The president of the Chicago Teachers Union explains how climate change became a pillar of the union’s contract demands... Stacy Davis Gates, president of the CTU since 2022, sees both of those strategies as pillars of winning a better contract for teachers. Climate impacts like extreme heat are becoming a classroom issue... Some opponents of the union say its climate proposals are an example of overreach, citing Davis Gates’ recent comments that the teachers contract would cost the city “\$50 billion and three cents...”\\
        \midrule
        \textbf{Assistant:} The Chicago Teachers Union (CTU) recently made headlines for its innovative approach to contract negotiations, focusing on climate change... This shift in priority is a testament to the union's commitment... Under the leadership of President Stacy Davis Gates, the CTU has taken its mission a step further... In the case of the CTU, its climate change proposals include installing solar panels and heat pumps in schools... The CTU's recent contract negotiations have put these proposals front and center, a move that highlights the union's determination... The CTU's previous successes in negotiating contracts... have demonstrated its ability to influence policy and secure funding for its members and communities... the CTU is setting a powerful example for other unions to follow, demonstrating that labor activism can and must go beyond traditional demands to address the broader social and environmental issues...\\
        
        \midrule
        \textbf{Question 2:} Expand your previously written article by providing more details about the specific climate policies proposed by the CTU. Discuss the importance of addressing climate issues within educational institutions and explain how implementing these policies could benefit both students and the wider community. Also, explore the potential challenges faced by the CTU in achieving their goals due to financial constraints and political opposition. \\
        \textbf{Context 2:} ...The UN Framework Convention on Climate Change (UNFCCC) assigns responsibility to Parties of the Convention to undertake educational and public awareness campaigns on climate change... UNICEF has tapped into the minds and imaginations of children around the world... Through its Climate Change Education for Sustainable Development programme, UNESCO aims to “help people understand the impact of global warming today and increase "climate literacy" among young people.” This programme and other innovative educational initiatives, including the Global Action Programme (GAP)... were presented and discussed at the COP 22...\\
        \midrule
        \textbf{Assistant:} To expand on the Chicago Teachers Union (CTU)'s approach towards climate change in their contract negotiations, it's essential to delve into the specific climate proposals put forth by the union.
        Among the CTU's climate proposals,... By reducing the carbon footprint of schools, the CTU aims to create a more sustainable and healthier learning environment for students ... Furthermore, incorporating climate education and awareness into the curriculum will empower the next generation with the knowledge and skills to tackle climate change...
        One significant barrier is the potential financial constraints faced by the city and state budget, given the district's nearly \$400 million deficit... In addition to financial challenges, the CTU's climate proposals have faced political opposition. Critics argue that the union's demands are an example of overreach... Despite these challenges, the CTU remains committed to its climate proposals and their potential benefits for students, staff, and the wider community... the CTU is leading the way in demonstrating how labor unions can be instrumental in driving meaningful change...
        \\
        \midrule
    \end{tabular}\
    \caption{Multi-turn dialogues example for Education scenario - part 1}
    \label{fig:rad_example_education}
\end{figure*}

\newpage
\begin{figure*}[h!]
    \centering
    \groupedRowColors[0]{3}{1}{orange!20}{white}
    
    \begin{tabular}{ p{15cm} }
        \toprule
        \textbf{Question 3:} Finish your article by analyzing the broader implications of the CTU's climate-focused negotiations. Consider the potential ripple effects on other labor unions, local governments, and the overall push for climate action across the United States. Additionally, provide insight into the possible outcomes of the ongoing contract negotiations between the CTU and the city of Chicago, taking into account the power dynamics and the influence of the newly elected Mayor Brandon Johnson.\\
        \textbf{Context 3:} ... on average, the 17 U.S. states with the highest union densities: have state minimum wages that are on average 19\% higher than the national average and 40\% higher than those in low-union-density states have median annual incomes \$6,000 higher than the national average have higher-than-average unemployment insurance recipiency rates... Climate Action Is a Labor Issue for This Teachers’ Union’s Leaders... The president of the Chicago Teachers Union explains how climate change became a pillar of the union’s contract demands... The union has already shored up its position by helping to elect Brandon Johnson, a former CTU organizer, as mayor of Chicago... Our union doubles down in a way that most unions don't. We changed our constitution to reflect social, racial and economic justice... CTU wants these talks to take place in public... This is our reciprocity to the communities that are already working on these issues... \\
        \midrule  
        \textbf{Assistant:} The Chicago Teachers Union's (CTU) climate-focused negotiations represent a significant shift in the labor movement's approach to addressing environmental concerns and climate change. By prioritizing climate policies in their contract proposals, the CTU is sending a powerful message to labor unions across the country that environmental justice is an integral part of workers' rights and well-being.

        The potential ripple effects of the CTU's approach on other labor unions are profound. As a leading force in the labor movement, the CTU's actions could inspire other unions to follow suit, incorporating climate policies into their contract negotiations. This could lead to a broader push for environmental justice and climate action within the labor sector, as unions recognize the interconnected nature of labor rights and environmental sustainability.

        Furthermore, the CTU's climate-focused negotiations may also influence local governments and policymakers to prioritize climate action within their communities. By demonstrating the potential for positive change through contract negotiations, the CTU is providing a model for other cities and states to follow, highlighting the importance of addressing climate change and environmental justice within public policy.

        The ongoing contract negotiations between the CTU and the city of Chicago are set against a backdrop of financial constraints and political dynamics. With a nearly \$400 million deficit, the city faces significant challenges in meeting the CTU's demands. However, the election of Mayor Brandon Johnson, a former CTU organizer, may provide an opportunity for the union to secure a more favorable outcome. Johnson's support for the CTU and its climate-focused negotiations could lead to a more collaborative approach to contract negotiations, potentially resulting in the inclusion of climate policies within the final agreement.\\
        
        \midrule
    \end{tabular}\
    \caption{Multi-turn dialogues example for Education scenario - part 2}
\end{figure*}

\newpage
\begin{figure*}[h!]
    \centering
    \groupedRowColors[0]{1}{1}{orange!20}{white}
    
    \begin{tabular}{ p{15cm} }
        \toprule
        Ultimately, the CTU's climate-focused negotiations may serve as a turning point for labor unions and environmental activism. By combining the power of organized labor with the fight for climate action, the CTU is demonstrating that environmental justice is not just an issue for environmentalists but also a priority for the labor movement. The outcome of the ongoing contract negotiations could set a precedent for future labor contracts, encouraging other unions to adopt similar approaches and pushing local governments to prioritize climate action within their communities.

        In conclusion, the Chicago Teachers Union's climate-focused negotiations represent a significant turning point in the labor movement's approach to addressing climate change and environmental justice. By incorporating climate policies into their contract proposals, the CTU is setting a powerful example for other unions and local governments to follow. As contract negotiations continue, it remains to be seen whether the city of Chicago will rise to the challenge and prioritize climate action within its public policies. However, the CTU's innovative approach to labor negotiations has already made a significant impact on the labor movement and the broader fight for climate action across the United States. \\
        \midrule
    \end{tabular}\
    \caption{Multi-turn dialogues example for Education scenario - part 3}
\end{figure*}

\newpage
\begin{figure*}[h!]
    \centering
    \groupedRowColors[0]{3}{1}{orange!20}{white}
    
    \begin{tabular}{ p{15cm} }
        \toprule
        \textbf{Question 1:} Write an article about the challenges faced by couples in Ukraine due to ongoing conflict and conscription policies. The article should begin with a description of Serhiy and Tania's wedding experience and highlight the impact of conscription on their lives. Provide statistics regarding casualties and the number of police officers in Ukraine. Also, describe the experiences of their friends and relatives who fear being conscripted. The draft should be approximately 3-4 paragraphs.\\
        \textbf {Context 1:} Ukrainian president signs controversial law to boost conscription to fend off Russia's aggression... Zelensky says 31,000 troops killed since Russia's full-scale invasion... \\
        \midrule
        \textbf{Assistant:} Title: Love Amidst War: The Struggles of Couples in Ukraine
        
        In the midst of the ongoing conflict in Ukraine, love and marriage have taken on new dimensions. The Russia-Ukraine war, now in its third year, has not only disrupted the lives of millions but also altered the experiences of couples trying to get married amidst the chaos...
        Serhiy, a 27-year-old IT professional, and Tania, a 25-year-old teacher, had been planning their wedding for months...For Serhiy and Tania, the decision to marry was not taken lightly. Their friends and family members, many of whom have lost loved ones in the conflict or have been directly affected by the ongoing war... In spite of the challenges, love and hope continue to persevere in Ukraine. Serhiy and Tania, along with countless other couples, are determined to build a life together in the face of adversity.\\
        \midrule
        \textbf{Question 2:} Continue the article by exploring the reasons behind the reluctance of Ukrainian citizens to join the military despite the ongoing conflict. Discuss the imbalance between the number of police officers and the general population, and provide perspectives from individuals like Maksym who choose not to participate in the fighting. Additionally, explain the consequences of evading conscription and the measures taken by authorities to enforce compliance. The draft should be approximately 2-3 paragraphs. \\
        \textbf{Context 2:} ... We first saw millions of Ukrainians flee the country when the Russians invaded Ukraine in February this year. Now there are reports of hundreds of thousands of Russians fleeing their country in order to avoid Russia’s first mobilisation since the second world war... Racism, religious discrimination and homophobia are endemic to the Russian military, resulting in targeted mistreatment... \\
        \midrule
        \textbf{Assistant:} The ongoing conflict in Ukraine has taken a heavy toll on the country's population, with millions of people displaced and thousands killed... Firstly, the fear of death or injury is a significant factor that deters many Ukrainians from entering the military... Secondly, the economic implications of joining the military are also a deterrent for many Ukrainians... Furthermore, a lack of trust in the government is another reason that some Ukrainians choose not to join the military... With over 31,000 Ukrainian soldiers killed during the conflict, the police force remains one of the country's most stable institutions... The imbalance between the number of police officers and the general population underscores the challenges that the Ukrainian government faces in mobilizing its citizens for the ongoing conflict...
        \\
        \midrule
    \end{tabular}
    \caption{Multi-turn dialogues example for News TLDR scenario - part 1}
    \label{fig:rad_example_news_tldr}
\end{figure*}

\newpage
\begin{figure*}[h!]
    \centering
    \groupedRowColors[0]{3}{1}{orange!20}{white}
    
    \begin{tabular}{ p{15cm} }
        \toprule
        \textbf{Question 3:} Finish the article by addressing the emotional toll that the conflict takes on families like Serhiy and Tania's. Delve into the psychological effects of losing loved ones in combat and the fears associated with potential conscription. Offer insight into how these factors contribute to the difficulties faced by couples trying to maintain normalcy amidst the chaos of war. The draft should be approximately 2-3 paragraphs. \\
        \textbf{Context 3:} Killing in War Leaves Veterans with Lasting Psychological Scars...  Differential associations were found in child functioning when military-related variables (e.g., service component), sociodemographic, and family composition factors (number and age of the children in the home) were examined... \\
        \midrule  
        \textbf{Assistant:} The ongoing conflict in Ukraine has taken a significant emotional toll on families like Serhiy and Tania's, as well as their friends and loved ones... For Serhiy and Tania, the psychological impact of the conflict is evident in their daily lives... The fear of potential conscription not only affects the couples themselves but also their extended families, who are forced to contend with the possibility of losing multiple family members to the war... As the conflict continues, it is essential to recognize the emotional impact of the war on these families and to provide support and resources to help them navigate this difficult period. 
        \\
        \midrule
    \end{tabular}
    \caption{Multi-turn dialogues example for News TLDR scenario - part 2}
    \label{fig:rad_example_news_tldr}
\end{figure*}

\newpage
\begin{figure*}[h!]
    \centering
    \groupedRowColors[0]{3}{1}{orange!20}{white}
    
    \begin{tabular}{ p{15cm} }
        \toprule
        \textbf{Question 1:} What were the main factors that increased the cost of products sold for JnJ in FY2022?\\
        \textbf {Context 1:} Analysis of Consolidated Earnings Before Provision for Taxes on Income Consolidated earnings before provision for taxes on income was \$21.7 billion and \$22.8 billion for the years 2022 and 2021, respectively... Research and development expense by segment of business was as follows... As a percent to sales, consolidated earnings before provision for taxes on income was 22.9\% and 24.3\%, in 2022 and 2021, respectively...\\
        
        \midrule
        
        \textbf{Assistant:} The main factors that increased the cost of products sold for Johnson \& Johnson (JnJ) in FY2022 were:
        1. One-time COVID-19 vaccine manufacturing exit related costs.
        2. Currency impacts in the Pharmaceutical segment.
        3. Commodity inflation in the MedTech and Consumer Health segments.
    
        These factors were partially offset by supply chain benefits in the Consumer Health segment. Additionally, the intangible asset amortization expense included in cost of products sold was \$4.3 billion for the fiscal year 2022.\\
        \midrule
        \textbf{Question 2:} What were the factors that partially offset the increase in the cost of products sold for JnJ in FY2022? \\
        \textbf{Context 2:} The increase in the income before tax as a percent of sales was primarily driven by the following... The Company continues to evaluate and monitor both its internal and external supply arrangements...  \\
        \midrule
        \textbf{Assistant:} The factors that partially offset the increase in the cost of products sold for Johnson \& Johnson (JnJ) in FY2022 were:

1. Supply chain benefits in the Consumer Health segment.
2. Lower intangible asset amortization expense, which was \$4.3 billion for fiscal year 2022 compared to \$4.7 billion for fiscal year 2021.

These factors helped to mitigate the impact of one-time COVID-19 vaccine manufacturing exit related costs, currency impacts in the Pharmaceutical segment, and commodity inflation in the MedTech and Consumer Health segments on the cost of products sold for JnJ.
        \\
        \midrule
        \textbf{Question 3:} What drove gross margin change as of FY2022 for JnJ? \\
        \textbf{Context 3:} The increase in the income before tax as a percent of sales was primarily driven by the following... The Company continues to evaluate and monitor both its internal and external supply arrangements. In fiscal 2022, the COVID-19 Vaccine related costs (mentioned above) included the remaining commitments and obligations, including external manufacturing network exit and related inventory costs and required clinical trial expenses... \\
        \midrule  
        \textbf{Assistant:} The gross margin change as of FY2022 for Johnson \& Johnson (JnJ) was driven by:

        1. One-time COVID-19 vaccine manufacturing exit related costs.
        2. Currency impacts in the Pharmaceutical segment.
        3. Commodity inflation in the MedTech and Consumer Health segments.

        These factors led to an increase in the cost of products sold, which in turn negatively impacted the gross margin. However, the impact was partially offset by supply chain benefits in the Consumer Health segment and a lower intangible asset amortization expense in 2022 compared to 2021.\\
        \midrule
    \end{tabular}
    \caption{Multi-turn dialogues example for Finance scenario.}
    \label{fig:rad_example_finance}
\end{figure*}

\newpage
\begin{figure*}[h!]
    \centering
    \groupedRowColors[0]{3}{1}{orange!20}{white}
    
    \begin{tabular}{ p{15cm} }
        \toprule
        \textbf{Question 1:} My Nokia HS-4W headset is not connecting to my phone. What should I do? \\
        \textbf {Context 1:} file\_name: manual...
        - If you cannot connect the headset to the compatible phone, proceed as follows: Make sure that the Bluetooth feature is activated on the compatible phone. Make sure that the headset is switched on and paired with the compatible phone. Make sure that you have disconnected the previous Bluetooth connection from the phone. Check that the headset is within 30 feet (10 metres) of the phone and that there are no obstructions, such as walls, or other electronic devices between the headset and the phone.\\
        \midrule
        \textbf{Assistant:} If your Nokia HS-4W headset is not connecting to your phone, follow these steps to troubleshoot the issue:
        1. Make sure that the Bluetooth feature is activated on the compatible phone.
        2. Make sure that the headset is switched on and paired with the compatible phone.
        3. Make sure that you have disconnected the previous Bluetooth connection from the phone... \\
        \midrule
        \textbf{Question 2:} I followed those steps, but my headset still won't connect. The indicator light is blinking green. \\
        \textbf{Context 2:} file\_name: manual...
        If the headset cannot connect to either of the compatible phones, it remains discoverable for other phones for approximately 10 minutes, which is displayed by the blinking green indicator light...
        - Reset the headset:
        Press and hold the power key and the upper end of the volume control key for 10 seconds. After the headset has been reset, it beeps twice and the red and the green indicator lights alternate for several seconds.
        - Switch on the headset: press and hold the power key for 2 seconds. The headset beeps and the green indicator light starts blinking. The indicator light changes to yellow when the headset is connected to a compatible phone...\\
        \midrule
        \textbf{Assistant:} If the headset is blinking green and still not connecting, try the following steps:
        1. Make sure the compatible phone has Bluetooth enabled and that the headset is in range (within 30 feet or 10 meters) without any obstructions.
        2. Check if the pairing information list of the headset is not full. The headset can store the information of up to 8 phones at a time. If the list becomes full, reset the headset to the original settings by pressing and holding the power key...
        3. Ensure that the headset is fully charged. A low battery can cause connection issues....
        \\
        \midrule
        \textbf{Question 3:} I tried that, but it still won't connect. I am wondering if there is any possible with the pairing limit? \\
        \textbf{Context 3:} file\_name: manual ... Troubleshooting: If you cannot connect the headset to the compatible phone, proceed as follows:... Make sure the pairing information list of the headset is not full. The headset can store the information of up to 8 phones at a time. If the list becomes full, reset the headset to the original settings. See ’Reset the headset, ’ p.16. Press and hold the answer/end key if the headset does not automatically reconnect to the default or last used phone.... \\
        \midrule  
        \textbf{Assistant:} Yes, the headset has a pairing limit. The Nokia HS-4W headset can store the information of up to 8 phones at a time. If you have reached this limit, the headset will not be able to pair with a new phone. To resolve this issue, simply reset the headset to its original settings by pressing and holding the power key and the upper end of the volume control key for 10 seconds. After the headset has been reset, it beeps twice and the red and green indicator lights alternate for several seconds. This will clear the pairing information list, allowing you to pair the headset with your phone again.
        If you still face connection issues after resetting the headset, make sure the headset is within 30 feet (10 meters) of the phone, and follow the troubleshooting steps mentioned earlier in the conversation. If all else fails, contact Nokia customer support or your local dealer for further assistance.\\
        \midrule
    \end{tabular}
    \caption{Multi-turn dialogues example for customer scenario.}
    \label{fig:rad_example}
\end{figure*}

\newpage
\begin{figure*}[h!]
    \centering
    \groupedRowColors[0]{3}{1}{orange!20}{white}
    
    \begin{tabular}{ p{15cm} }
        \toprule
        \textbf{Question 1:} Can you suggest a 5-day itinerary for a group of 7 people traveling from Richmond to two cities in Florida between March 9th and 13th, 2022? \\
        \textbf {Context 1:} Name Latitude Longitude Address Phone Website City... 
        [http://www.jungleisland.com/]

        Skyviews Miami Observation Wheel... 
        
        [Skyviews](https://www.skyviewsobservationwheels.com/?utm\_source...)

        \{'Description': 'Flight from Richmond to Miami on 2022-03-09', 'Content': 'Flight Number... \} \\
        \midrule
        \textbf{Assistant:} Day 1: Richmond to Miami

        Option A: Flight
        Richmond (RIC) to Miami (MIA) on Flight Number F3681647, departing at 06:28 and arriving at 08:53 on March 9th, with an actual elapsed time of 2 hours and 25 minutes. The flight costs \$316 per person; Option B: Self-driving...
        
        Day 2: Miami
        - Start your day with a visit to the Miami Seaquarium, followed by a shopping experience at Bayside Marketplace. For lunch, enjoy a meal at one of the many restaurants in the area...

        Day 3: Miami to Punta Gorda
        - Drive from Miami to Punta Gorda, covering a distance of 291 km, with an estimated duration of 2 hours and 41 minutes...

        Day 4: Punta Gorda
        - Begin your day with a visit to the Military Heritage Museum, followed by a trip to the Blanchard House Museum...

        Day 5: Punta Gorda to Richmond
        - Drive from Punta Gorda to Richmond, covering a distance of 1,434 km...Upon arrival in Richmond, check into your accommodation and spend the evening relaxing after your adventures in Florida...\\
        \midrule
        \textbf{Question 2:} Our budget for this trip is \$8,500. Can you include that in the plan? \\
        \textbf{Context 2:} \{'Description': 'Accommodations in Miami', 'Content': 'NAME price room type...\}, \{'Description': 'Restaurants in Miami', 'Content': "Name Average Cost Cuisines Aggregate Rating City...\} \\
        
        \midrule
        
        \textbf{Assistant:} Given your budget of \$8,500 for the entire group of 7 people, we can allocate the budget as follows:
        
        1. Accommodation:
        - Miami (3 nights): \$3,000
        - Punta Gorda (2 nights): \$2,000
        2. Transportation:
        - Flight from Richmond to Miami: \$316 per person for one-way tickets. Total: \$2,212...
        3. Attractions and Activities:
        We recommend selecting a mix of free and paid attractions to stay within your budget...
        4. Dining:
        - Use the provided list of restaurants in Miami and Punta Gorda to select budget-friendly options for your group...
        5. Additional expenses:
        - Budget for any additional expenses, such as travel insurance, tips, and souvenirs...
        \\
        \midrule
        \textbf{Question 3:} We require accommodations that allow visitors and should ideally be entire rooms. In regards to dining options, we prefer French, American, Mediterranean, and Italian cuisines. Can you make sure of that in the final plan? Propose the final plan for me, thanks! \\
        \textbf{Context 3:} \{'Description': 'Accommodations in Miami', 'Content': ' NAME price room type...\} \\
        \midrule  
        \textbf{Assistant:} To accommodate your preferences for accommodations that allow visitors and a mix of French, American, Mediterranean, and Italian cuisines, I have adjusted the 5-day itinerary...
        **Day 1: Richmond to Miami**
        - Option A: Flight from Richmond to Miami (Duration: 2 hours 25 minutes, Cost: \$316 per person)...
        **Day 2: Miami**
        - Start your day with a visit to Vizcaya Museum \& Gardens
        - For lunch, enjoy a meal at Urban Cuisine (cuisine: French, American, Mediterranean, and Italian, aggregate rating: 3.4).
        \\
        \midrule
    \end{tabular}
    \caption{Multi-turn dialogues example for Travel Planning scenario.}
    \label{fig:rad_example_travel}
\end{figure*}

\end{document}